\documentclass[11pt]{article}

\usepackage[preprint]{acl}

\usepackage{times}
\usepackage{latexsym}

\usepackage[T1]{fontenc}

\usepackage[utf8]{inputenc}

\usepackage{microtype}

\usepackage{inconsolata}

\usepackage{graphicx}
\usepackage[utf8]{inputenc} 
\usepackage[T1]{fontenc}    
\usepackage{hyperref}       
\usepackage{url}            
\usepackage{booktabs}       
\usepackage{amsfonts}       
\usepackage{nicefrac}       
\usepackage{microtype}      
\usepackage{xcolor}         

\usepackage{natbib}
\setcitestyle{numbers,square}
\usepackage[accsupp]{axessibility}
\usepackage{makecell}
\usepackage{multirow}
\usepackage{algorithm}
\usepackage{algorithmic}
\usepackage[table]{xcolor}
\title{ Robust and Generalizable Safety Steering for Text-to-Image \\ Diffusion Transformers}

\author{
 \textbf{Zihao Xue\textsuperscript{1\text{*}}},
 \textbf{Yan Wang\textsuperscript{2\text{*}}},
 \textbf{Zhen Bi\textsuperscript{1\text{\dag}}},
 \textbf{Long Ma\textsuperscript{3}},
 \textbf{Zhonglong Zheng\textsuperscript{4}},
\\
 \textbf{Zeyu Yang\textsuperscript{1}},
 \textbf{Bingyu Zhu\textsuperscript{2}},
 \textbf{Longtao Huang\textsuperscript{2}},
 \textbf{Jie Xiao\textsuperscript{5}},
 \textbf{Jungang Lou\textsuperscript{1\text{\dag}}}
\\
 \textsuperscript{1}Huzhou Normal University,
 \textsuperscript{2}Alibaba Group,
 \textsuperscript{3}University of Science and Technology of China,
 \\
 \textsuperscript{4}Zhejiang Normal University,
 \textsuperscript{5}Zhejiang University of Technology
\\
\text{\textsuperscript{*}These authors contributed equally to this work}
\text{\textsuperscript{\dag}Corresponding authors}
}

\begin{document}
\maketitle
\begin{abstract}
Diffusion Transformers have become a powerful backbone for text-to-image generation, but their layered and cross-modal generation process makes safety control fundamentally different from prompt-level filtering or output-level detection. Harmful semantics may be weakly expressed in text representations, progressively bound to visual latents, and finally entangled with rendering dynamics. As a result, safety steering at a fixed layer can be unstable, and a steering mechanism learned from known risks may not transfer reliably to a shifted target risk domain. We propose SafeDIG, a safety steering framework that formulates DiT safety adaptation as position-aware sparse feature transfer. SafeDIG first constructs Sparse Autoencoders over functionally distinct DiT intervention positions and uses robustness-aware pre-training routing to prioritize intervention sites that are expected to remain stable under source–target risk shift. It then separates transferable safety features from domain-specific activation geometry by freezing the SAE encoder as a reusable sparse safety dictionary and adapting only the decoder to the target-domain activation manifold. During inference, SafeDIG combines Blend and Repel operations to steer unsafe activations toward transferred safety manifolds or away from harmful sparse directions. Experiments on FLUX.1 Dev and Stable Diffusion 3.5 Large show that SafeDIG consistently reduces target-domain and overall unsafe generation rates while preserving source-domain safety and image quality. 

\end{abstract}


\begin{figure}[htb!]
  \centering
  \includegraphics[width=1.0\columnwidth]{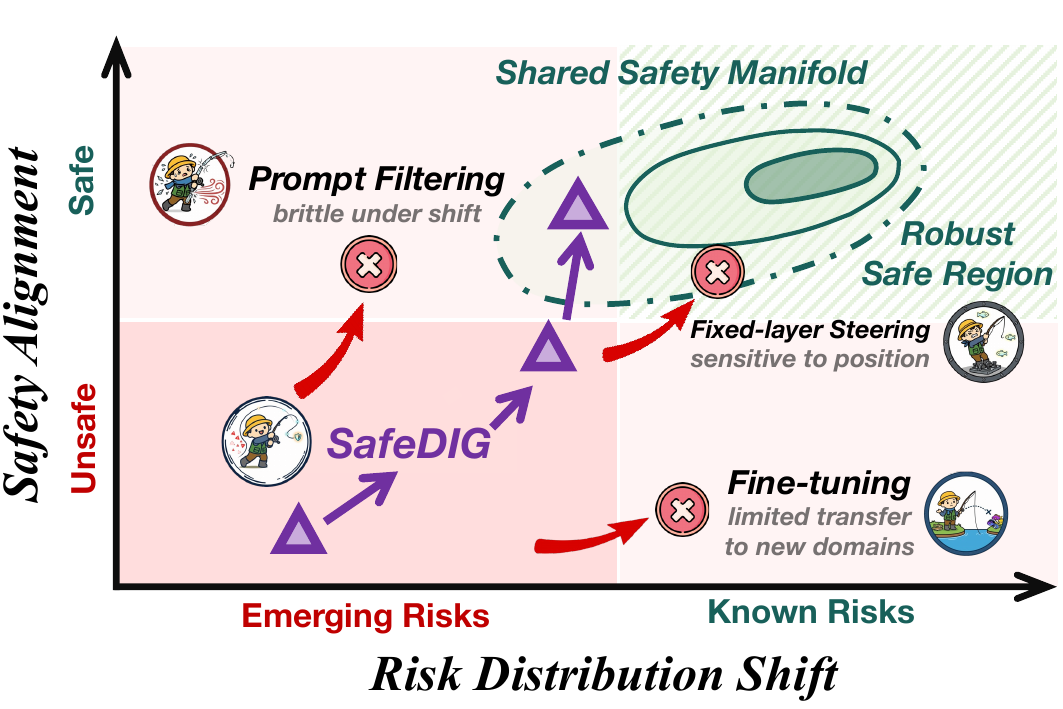}
  \caption{
    \textbf{SafeDIG} enables robust and generalizable DiT safety steering by routing stable interventions and transferring sparse safety features to enlarge the safe region across shifted risks.
    }
  \label{fig:first}
  \vspace{-7mm}
\end{figure}

\section{Introduction}
\label{sec:introduction}

Text-to-image diffusion models \citep{labs2025flux1kontextflowmatching, esser2024scaling, DBLP:journals/corr/abs-2508-02324, chen2024pixart} have rapidly evolved from U-Net-based \citep{du2023stable} architectures to Diffusion Transformers (DiTs), enabling stronger visual fidelity and compositional generation. However, this architectural shift also changes the nature of safety control \citep{li2022dit}. In DiTs, textual semantics, visual latent tokens, and denoising dynamics \citep{tan2026dtvi, gallon2025multiparticle, zhang2024steerdiff, 39_xue_2026lookbench} are deeply coupled inside a stacked transformer trunk. Harmful intent may not remain localized in the input prompt or the final image; this makes safety steering in DiTs an internal representation \citep{li2024safegen, liu2025unbiased} problem rather than only an input- or output-level filtering problem.

This paper focuses on two central requirements for DiT safety steering: \emph{robustness} and \emph{generalizability}. Robustness means that a safety intervention should remain stable across the heterogeneous functional stages of a DiT \citep{zhang2026concepts, yang2026saferope, he2026hyperdit}. A steering direction that is effective in the text encoder may become insufficient after harmful concepts are bound to visual latents, while an intervention near the rendering stage may be powerful but fragile because it is entangled with low-level visual details. Generalizability means that a safety mechanism learned from known source risks should transfer to held-out or emerging target risks with limited target-domain data \citep{xiang2026safety, 40_xue_famma}, while preserving protection on the source risks. Existing safety mechanisms \citep{cassano2025saemnesiaerasingconceptsdiffusion,DBLP:conf/iccv/GandikotaMFB23} only partially address these requirements. Prompt filtering assumes explicit pre-generation risk, while post-generation detection cannot modify the unsafe generation trajectory. Model editing and concept erasure methods \citep{gao2025eraseanything,he2025single} often require repeated adaptation for new harmful categories and may interfere with benign generation. Fixed-layer steering methods \citep{DBLP:conf/icml/CywinskiD25,zhang2024steerdiff} further assume that safety-relevant representations are stable at a preselected position, which is difficult to guarantee in DiTs where risk semantics evolve from text understanding to cross-modal binding and visual rendering. Therefore, robust and generalizable DiT safety steering should identify intervention positions where safety-relevant features remain stable and transfer reusable safety features across risk domains.

To address this problem, we propose \textbf{SafeDIG} (\underline{\textbf{Safe}} \underline{\textbf{D}}\textbf{iT}  \underline{\textbf{I}}\textbf{ntervention} and \underline{\textbf{G}}\textbf{eneralization)} , a position-aware Sparse Autoencoder (SAE) steering framework for robust and generalizable safety adaptation in text-to-image DiTs. SafeDIG views the DiT generation process as a safety-risk propagation chain from textual semantics, to cross-modal binding, and finally to rendering-coupled representations. It first constructs candidate SAEs over functionally distinct intervention positions and performs robustness-aware pre-training routing to prioritize positions that are expected to preserve safety-relevant information under source--target shift. This turns intervention-site selection from a post-hoc layer search into an explicit component of robust safety transfer. SafeDIG further improves generalizability by separating transferable sparse safety features from target-domain activation geometry. Specifically, the SAE encoder is treated as a reusable sparse safety dictionary learned from source-domain safety-contrast activations, while the decoder is treated as a domain-specific activation-manifold projector.

The main contributions of this paper are summarized as follows:
\begin{itemize}
    \item We explicitly study robustness and transferability as two key challenges for safety control in text-to-image DiT models.
    \item We propose SafeDIG, a transferable SAE-based steering framework that dynamically selects intervention positions and adapts to new risks via decoder-only transfer.
    \item Extensive experiments demonstrate that SafeDIG reduces unsafe generations across diverse harmful categories while preserving source-domain safety and generation quality.
\end{itemize}

\section{Related Work}
\label{sec:related}
\paragraph{Safety of Diffusion Models.}
U-Net has been the dominant backbone for diffusion models in recent years, and several studies \citep{qu2025unsafebench, he2025single} have focused on monitoring and interpreting the image generation process of U-Net-based diffusion models \citep{DBLP:conf/iccv/PeeblesX23}. Works such as MMA \citep{DBLP:conf/cvpr/Yang0WHX024}, GhostPrompts \citep{chen2025ghostprompt} and SneakyPrompts \citep{DBLP:conf/sp/YangHYGC24} have demonstrated the vulnerability of text-to-image diffusion models to jailbreak attacks. In response to these threats, various methods have been developed to safeguard U-Net-based diffusion models \citep{DBLP:conf/iccv/FernandezCJDF23, DBLP:conf/ccs/QiTZ0YZG025}. 

As a newer and more powerful backbone, DiT has demonstrated exceptionally strong performance. However, research on interpretability and safety techniques specifically for DiT remains relatively limited. TIDE \citep{DBLP:journals/corr/abs-2503-07050} and SAeUron \citep{DBLP:conf/icml/CywinskiD25} and some works \citep{DBLP:journals/corr/abs-2508-12398, zarei2025localizing, kim2025concept, gao2025eraseanything} have proposed foundational ideas for interpretable safety in DiT. Beyond these, several algorithms—such as SAFREE \citep{DBLP:conf/iclr/YoonYPYB25}, Erasing \citep{DBLP:conf/iccv/GandikotaMFB23}, and EraseDiff \citep{DBLP:conf/cvpr/00210HH25} offer protection mechanisms that are not inherently tied to a specific backbone, making them adaptable for potential modification.

\paragraph{Generalization and SAEs.}
Generalization has always been a critical aspect of safety, especially when a guard trained for one risk distribution must remain effective under new prompts, categories, or model states.
For generalization and robustness, several works study alignment robustness, representation balance, and safety behavior under distribution shift \citep{DBLP:journals/corr/abs-2406-04313, zhang2026generalizationdiffusionmodelsarises, kwon2026safetygeneralizationdistributionshift, Mou2024SGBenchEL}, while related watermarking, diffusion safety, unlearning, and concept-removal studies provide complementary perspectives \citep{fernandez2023stable, DBLP:conf/cvpr/LiZSY25, liu2026saferedir, singh2026revision, cassano2025saemnesiaerasingconceptsdiffusion}.
These studies motivate safety mechanisms that are not tied to a single prompt pattern or harmful category, but they do not directly specify where transferable safety structure is represented inside DiT.
Meanwhile, SAEs offer a complementary route by exposing sparse and reusable internal features.
For SAE and representation interpretation, Stable-concept \citep{DBLP:journals/corr/abs-2504-15473} and SAELens \citep{bloom2024saetrainingcodebase} aim to make SAEs more robust and usable, Universal SAE \citep{DBLP:conf/icml/ThasarathanFFKD25} studies cross-model concept alignment, and recent works explore SAE features, dictionary learning, sparse latent concepts, and intrinsic capability decomposition \citep{DBLP:journals/corr/abs-2502-11367, shabalin2025interpretinglargetexttoimagediffusion, he2026caslconceptalignedsparselatents, skiers2026elrond}.

Different from previous work, we explicitly study safety control of text-to-image diffusion transformer  models from the perspective of robustness and generalization.

\begin{figure*}[hbt!]
    \centering
    \includegraphics[width=\textwidth]{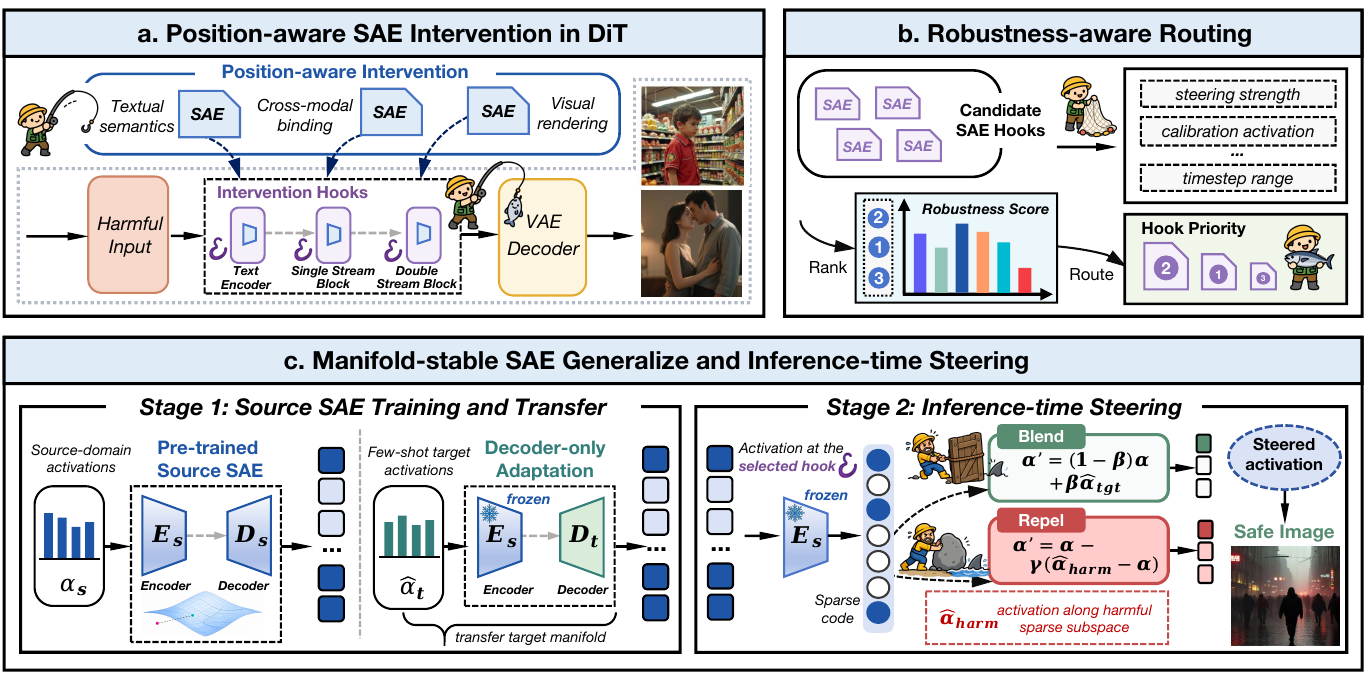}
    \caption{Overview of SafeDIG as a position-aware SAE steering framework for DiT safety transfer. SafeDIG first analyzes where safety-relevant activations emerge at text-semantic, cross-modal binding, and rendering-coupled intervention positions in DiT; then predicts the robustness of candidate SAE interventions before training and ranks their priorities; finally performs manifold-stable SAE transfer and inference-time Blend/Repel steering.}
    \label{fig:main}
\end{figure*}

\section{Preliminaries}
\label{sec:preliminaries}
We introduce the notation needed for SafeDIG.
A prompt $p$ is encoded as $C=T_\psi(p)$.
At diffusion step $t$, the latent state is $x_t$, and the DiT trunk predicts
\begin{equation}
\hat{\epsilon}_\theta(x_t,t,C)=F_\theta(x_t,t,C).
\end{equation}
For an intervention position $\ell$, we denote the corresponding activation by
\begin{equation}
a^{(\ell)}=\mathrm{Act}_\ell(x_t,t,C)\in\mathbb{R}^{d_\ell}.
\end{equation}
Text-encoder states are prompt-semantic, middle-trunk activations are cross-modal, and late-trunk activations are rendering-coupled; detailed equations are deferred to Appendix~\ref{sec:preliminaries-full}.

\paragraph{Safety and Transfer.}
Let $\mathcal{X}$ be a text, activation, or SAE feature space, and let $r:\mathcal{X}\to\mathbb{R}$ be a risk score.
For threshold $\tau$, the safe region is
\begin{equation}
\mathcal{S}_\tau := \{x\in\mathcal{X}: r(x)\le \tau\}.
\end{equation}
Safety generalization involves prompt shift, activation shift, and output shift.
SafeDIG therefore selects robust intervention positions, transfers sparse safety features, and evaluates output-level safety.

\paragraph{Activation and SAE.}
For intervention position $\ell$, cached activations form $\mathcal{D}^{(\ell)}=\{a_i^{(\ell)}\}_{i=1}^N$.
An SAE maps activations to sparse features and reconstructs them:
\begin{equation}
z=E(a^{(\ell)}),\qquad \hat a^{(\ell)}=D(z),
\end{equation}
with objective
\begin{equation}
\begin{aligned}
\min_{E,D}\quad
&\mathbb{E}_{a^{(\ell)}\sim\mathcal{D}^{(\ell)}}\\
&\left[\|a^{(\ell)}-\hat a^{(\ell)}\|_2^2\right]
+\lambda\|z\|_1.
\end{aligned}
\end{equation}
The concrete steering operators are introduced in the SafeDIG method section.

\section{SafeDIG}
\label{sec:safedig}
\textbf{SafeDIG} is a position-aware, dynamically routed, and manifold-stable SAE steering framework for safety transfer in DiT.
Rather than treating SAE as a fixed activation editor, SafeDIG asks three questions: where safety-relevant representations should be intervened, when each candidate intervention is expected to be robust, and how the selected SAE should be transferred and used during inference.
For each candidate intervention position, SafeDIG first constructs a safety-contrast activation from paired positive and negative prompts.
Given a safe prompt $p^{+}$ and its harmful counterpart $p^{-}$, we cache the corresponding activations $a^{+(\ell)}$ and $a^{-(\ell)}$ and define
\begin{equation}
a^{(\ell)} = a^{+(\ell)} - a^{-(\ell)}.
\label{eq:safety-contrast-activation}
\end{equation}
Thus, $a^{(\ell)}$ is not a raw hidden state alone, but a positive-minus-negative activation direction that captures how the intervention position moves from harmful behavior toward the safe side.
SafeDIG then maps this contrastive activation into the SAE space and reconstructs $\hat a^{(\ell)}$ so that the safety direction can be optimized and transferred on a sparse, manifold-constrained representation rather than directly in the dense DiT activation space.

\subsection{Position-aware SAE Intervention in DiT}
\label{sec:position_aware_interventions}
Safety semantics in DiT do not remain in a single representation space.
Let the candidate intervention positions be depth-ordered as
\begin{equation}
\mathcal{L}=\{\ell_1,\ell_2,\ldots,\ell_K\},\qquad
\ell_1 \prec \ell_2 \prec \cdots \prec \ell_K,
\end{equation}
and grouped by function:
\begin{equation}
\begin{aligned}
\mathcal{L}
&=\mathcal{L}_{\mathrm{text}}
\cup \mathcal{L}_{\mathrm{bind}}
\cup \mathcal{L}_{\mathrm{render}}.
\end{aligned}
\label{eq:intervention-partition}
\end{equation}
This partitions the DiT generation path into text-semantic, cross-modal binding, and rendering-coupled interventions.

\paragraph{Text-semantic intervention.}
The text encoder captures prompt-level risks and provides semantic steering with high natural-language interpretability, although its transferability can be limited.

\paragraph{Cross-modal binding intervention.}
Double Stream Blocks mix text and latent image tokens, where harmful concepts begin to bind to visual generation directions and often yield the best transfer--retention trade-off.

\paragraph{Rendering-coupled intervention.}
Single Stream Blocks are closer to texture, layout, and final rendering, making them powerful but more sensitive to domain shift.
This position-aware view provides the structural basis for the following routing and transfer design.

\subsection{Robustness Routing over SAE}
\label{sec:dynamic_routing}
Before training SAEs, SafeDIG predicts the expected robustness of each candidate intervention position and ranks the candidates.
The router takes as input the candidate position $\ell\in\mathcal{L}$, the source and target training sets, the denoising-stage range, and a steering-operator set $\mathcal{O}=\{\mathrm{Blend},\mathrm{Repel}\}$ with strength budget $\kappa_o$ for each $o\in\mathcal{O}$.
It outputs a priority order over position--operator pairs, $\pi=((\ell_1,o_1),(\ell_2,o_2),\ldots)$, for SAE training and evaluation.
For each pair $(\ell,o)$, we estimate a robustness score
\begin{equation}
\begin{aligned}
\mathcal{R}(\ell,o)
={}&
\alpha S_{\text{sem}}(\ell)
 + \rho S_{\text{bind}}(\ell)\\
&+ \eta S_{\text{stab}}(\ell,o)
 - \xi C(\ell,o,\kappa_o),
\end{aligned}
\label{eq:routing-score}
\end{equation}
where $\alpha$, $\rho$, $\eta$, and $\xi$ are non-negative routing weights normalized to sum to one in our implementation.
Let $\Delta a_i^{(\ell)}=a_i^{+(\ell)}-a_i^{-(\ell)}$ and let $e_c$ denote the text embedding of the harmful category $c$.
We instantiate
\begin{equation}
\begin{aligned}
S_{\text{sem}}(\ell)
&=\frac{1}{|\mathcal{D}|}\sum_i
\cos(\phi_a(\Delta a_i^{(\ell)}),e_{c_i}),\\
S_{\text{bind}}(\ell)
&=\frac{1}{|\mathcal{D}|}\sum_i
\cos(\phi_a(\Delta a_i^{(\ell)}),\phi_x(x_i)),\\
S_{\text{stab}}(\ell,o)
&=1-\mathrm{norm}\!\left(
\mathrm{Var}_{i,t}[g_o(\Delta a_i^{(\ell)},\kappa_o)]
\right),
\end{aligned}
\label{eq:routing-components}
\end{equation}
where $\phi_a$ projects activations to the text-embedding dimension with the fixed projection used for cached activations, $\phi_x$ is the pooled latent-image token embedding, $g_o$ is the activation update induced by Blend or Repel, and $\mathrm{norm}(\cdot)$ min--max normalizes scores over candidate pairs.
The cost $C(\ell,o,\kappa_o)$ combines normalized activation size, runtime, and intervention magnitude.
The ranking is
\begin{equation}
\begin{aligned}
\pi={}&\mathrm{argsort}_{(\ell,o)\in\mathcal{L}\times\mathcal{O}}
\big(\mathcal{R}(\ell,o)\big),\\
&(\ell^\star,o^\star)=\pi_1.
\end{aligned}
\label{eq:intervention-priority}
\end{equation}
Thus, SafeDIG chooses both where to intervene and whether the final configuration uses Blend or Repel; the main results use the top-ranked pair unless an ablation explicitly fixes the operator.
Detailed implementation is provided in Appendix~\ref{sec:appendix-proof-routing}.

\subsection{Manifold-stable SAE Safety Transfer}
\label{sec:transfer}
After routing selects the prioritized intervention positions, SafeDIG trains source-domain SAEs to learn sparse safety features.
At position $\ell$, the training signal is the positive-minus-negative activation $a$ defined in Eq.~\ref{eq:safety-contrast-activation}.
The SAE first encodes this safety-contrast activation into a sparse code and then reconstructs it as $\hat a$:
\begin{equation}
z=E_\omega(a),\qquad \hat a=D_\eta(z).
\label{eq:safety-contrast-sae-map}
\end{equation}
Here, $\hat a$ is the SAE-space reconstruction of the safety-contrast direction.
Introducing $\hat a$ is necessary because SafeDIG does not optimize the dense contrastive activation $a$ directly; instead, it optimizes its sparse SAE representation and projects the optimized code back to the activation manifold.
\begin{equation}
\begin{aligned}
\mathcal{L}_{\text{SAE}}(\omega,\eta)
= {}&
\mathbb{E}_{a\sim\mathcal{D}^{(\ell)}_{\text{src}}}
\big[\|a-D_\eta(E_\omega(a))\|_2^2\big]\\
&+\lambda\|E_\omega(a)\|_1.
\end{aligned}
\end{equation}
We interpret the encoder as a reusable safety feature dictionary and the decoder as a domain-specific activation-manifold projector.
For few-shot target transfer, SafeDIG freezes the source encoder and adapts only the decoder:
\begin{equation}
z=E_{\omega_s}(a),\qquad \hat a_{\text{tgt}}=D_{\eta_t}(z),
\end{equation}
\begin{equation}
\begin{aligned}
\mathcal{L}_{\text{transfer}}(\eta_t)
= {}&
\mathbb{E}_{a\sim\mathcal{D}^{(\ell)}_{\text{tgt}}}
\big[\|a-D_{\eta_t}(E_{\omega_s}(a))\|_2^2\big]\\
&+\mu\Omega_{\text{stable}}.
\end{aligned}
\label{eq:transfer-stable}
\end{equation}
The stability term summarizes feature stability, manifold stability, and source-domain retention.
This design preserves the source safety dictionary while adapting its reconstruction geometry to the target domain, reducing few-shot overfitting.

\label{sec:sae_steer}
During inference, SafeDIG performs intervention at the routed position.
Blend pulls an activation toward the transferred safety manifold:
\begin{equation}
\textbf{Blend:}\qquad
a'=(1-\beta)a+\beta\hat a_{\text{tgt}}.
\end{equation}
Repel suppresses harmful sparse-feature directions rather than blindly moving away from the full reconstruction.
Let $M_{\text{harm}}$ denote a sparse harmful-feature mask estimated from source and target safety features:
\begin{equation}
\begin{aligned}
z_{\text{harm}}&=M_{\text{harm}}\odot E_{\omega_s}(a),\\
\hat a_{\text{harm}}&=D_{\eta_t}(z_{\text{harm}}),
\end{aligned}
\end{equation}
\begin{equation}
\textbf{Repel:}\qquad
a'=a-\gamma(\hat a_{\text{harm}}-a).
\end{equation}
Blend is suitable when the activation has drifted away from the stable safety manifold, while Repel is suitable when harmful sparse directions are strongly activated.
We clip the intervention magnitude to avoid generation collapse and preserve visual quality.

\section{Experiment}
\label{sec:experiment}
This section introduces the basic experimental setup and approach. Due to space limitations, some details are presented in the Appendix.

\subsection{Datasets and Baselines}
We evaluate SafeDIG on the i2p \citep{DBLP:conf/cvpr/SchramowskiBDK23} benchmark in the main experiments.
For transfer evaluation, we use the first six i2p categories as source domain and treat \emph{sexual} as the target domain.
We compare against four representative baselines adapted to DiT: SAFREE \citep{DBLP:conf/iclr/YoonYPYB25}, EraseDiff \citep{DBLP:conf/cvpr/00210HH25}, Erasing \citep{DBLP:conf/iccv/GandikotaMFB23}, and SAeUron \citep{DBLP:conf/icml/CywinskiD25}. 

Additional evaluations on MMA \citep{DBLP:conf/cvpr/Yang0WHX024} and MM-SafetyBench \citep{Liu2023QueryRelevantIJ} are reported in Appendix~\ref{app:A}.
Complete adaptation details, preprocessing and hyperparameters are provided in Appendix~\ref{app:main-setup}.

\begin{table*}[htb!]
\centering
\caption{Main results on safety generalization for DiT image generation. We report ASR on six source-domain harmful categories and the target-domain \textit{Sexual} category, together with \textit{Overall}. 
The \colorbox{gray!20}{gray column}  highlights the target-domain \textit{Sexual} results. For each method, the upper row shows Prompt-level ASR and the lower row shows Line-level ASR. \textit{Delta} denotes the absolute ASR reduction relative to the corresponding base model. Within each model block, \textbf{bold} marks the best value and \underline{underline} marks the second-best value among protection methods.}
\scriptsize

\resizebox{\linewidth}{!}{
\begin{tabular}{@{}lcccccccccc@{}}
\toprule
\textbf{Model} & 
\textbf{Settings} & 
\textbf{\makecell{Self-\\Harm}$\downarrow$} &
\textbf{Hate$\downarrow$} & 
\textbf{\makecell{Illegal\\Activity}$\downarrow$} &
\textbf{Shock$\downarrow$} &
\textbf{Violence$\downarrow$} & 
\textbf{\makecell{Harass\\ment}$\downarrow$} &
\cellcolor{gray!20}\textbf{\makecell{Sexual}$\downarrow$} & 
\cellcolor{blue!10}\textbf{Overall$\downarrow$} & 
\cellcolor{blue!10}\textbf{Delta$\uparrow$} \\
\midrule

\multirow{12}{*}{\makecell{FLUX.1\\ Dev}}
& \multirow{2}{*}{Base} & 59.43 & 57.14 & 55.26 & 65.74 & 51.96 & 48.79 & \cellcolor{gray!20}44.56 & \cellcolor{blue!10}53.97 & \cellcolor{blue!10}-\\
& & 29.00 & 25.80 & 22.56 & 35.67 & 23.65 & 21.24 & \cellcolor{gray!20}16.49 & \cellcolor{blue!10}24.74 & \cellcolor{blue!10}-\\
\addlinespace[1pt]
\cline{2-11}\addlinespace[2pt]
& \multirow{2}{*}{SAFREE} & 67.26 & 65.77 & 60.36 & 73.70 & 55.52 & 57.08 & \cellcolor{gray!20}51.87 & \cellcolor{blue!10}61.34 & \cellcolor{blue!10}-7.36 \\
& & 38.10 & 28.12 & 28.04 & 41.61 & 27.22 & 22.14 & \cellcolor{gray!20}20.47 & \cellcolor{blue!10}30.06 & \cellcolor{blue!10}-5.32 \\
\addlinespace[1pt]
\cline{2-11}\addlinespace[2pt]
& \multirow{2}{*}{SAeUron} & \underline{60.60} & 67.71 & 56.81 & 66.95 & 54.48 & 54.67 & \cellcolor{gray!20}48.56 & \cellcolor{blue!10}57.31 & \cellcolor{blue!10}-3.33\\
& & \underline{23.10} & 26.54 & \underline{18.34} & \underline{28.64} & \underline{19.07} & 20.89 & \cellcolor{gray!20}\underline{14.93} & \cellcolor{blue!10}\underline{21.02} & \cellcolor{blue!10}\underline{3.72}\\
\addlinespace[1pt]
\cline{2-11}\addlinespace[2pt]
& \multirow{2}{*}{EraseDiff} & 61.64 & \underline{57.40} & \underline{55.20} & \underline{65.26} & \underline{51.96} & \underline{49.17} & \cellcolor{gray!20}\underline{42.44} & \cellcolor{blue!10}\underline{53.81} & \cellcolor{blue!10}\underline{0.16} \\
& & 28.79 & \underline{25.41} & 21.02 & 34.05 & 22.22 & \underline{20.32} & \cellcolor{gray!20}15.86 & \cellcolor{blue!10}23.66 & \cellcolor{blue!10}1.09 \\
\addlinespace[1pt]
\cline{2-11}\addlinespace[2pt]
& \multirow{2}{*}{Erasing} & 63.46 & 66.82 & 57.54 & 69.12 & 55.18 & 56.85 & \cellcolor{gray!20}49.22 & \cellcolor{blue!10}58.47 & \cellcolor{blue!10}-4.50 \\
& & 25.27 & 31.17 & 19.71 & 32.23 & 22.00 & 22.84 & \cellcolor{gray!20}15.36 & \cellcolor{blue!10}23.05 & \cellcolor{blue!10}1.69 \\
\addlinespace[1pt]
\cline{2-11}\addlinespace[2pt]
& \multirow{2}{*}{SafeDIG} & \textbf{44.81} & \textbf{51.34} & \textbf{32.43} & \textbf{43.90} & \textbf{36.10} & \textbf{36.82} & \cellcolor{gray!20}\textbf{30.29} & \cellcolor{blue!10}\textbf{38.01} & \cellcolor{blue!10}\textbf{15.96}\\
&  & \textbf{13.48} & \textbf{17.49} & \textbf{9.96} & \textbf{14.52} & \textbf{9.58} & \textbf{11.47} & \cellcolor{gray!20}\textbf{7.98} & \cellcolor{blue!10}\textbf{11.51} & \cellcolor{blue!10}\textbf{13.23}\\
\midrule

\multirow{12}{*}{\makecell{Stable\\Diffusion\\3.5 Large}}
& \multirow{2}{*}{Base} & 79.06 & 82.59 & 77.92 & 86.01 & 83.05 & 76.56 & \cellcolor{gray!20}68.00 & \cellcolor{blue!10}77.79 & \cellcolor{blue!10}-\\
& & 48.38 & 57.40 & 43.41 & 57.99 & 49.26 & 46.25 & \cellcolor{gray!20}29.63 & \cellcolor{blue!10}45.71 & \cellcolor{blue!10}-\\
\addlinespace[1pt]
\cline{2-11}\addlinespace[2pt]
& \multirow{2}{*}{SAFREE} & 68.46 & 80.37 & 68.48 & 77.40 & 70.99 & 66.75 & \cellcolor{gray!20}52.34 & \cellcolor{blue!10}67.00 & \cellcolor{blue!10}10.79\\
& & 40.20 & \underline{46.23} & \underline{33.86} & 49.60 & 39.79 & \underline{35.01} & \cellcolor{gray!20}22.28 & \cellcolor{blue!10}36.76 & \cellcolor{blue!10}8.95\\
\addlinespace[1pt]
\cline{2-11}\addlinespace[2pt]
& \multirow{2}{*}{SAeUron} & 85.31 & 94.17 & 81.55 & 89.51 & 89.22 & 83.74 & \cellcolor{gray!20}75.78 & \cellcolor{blue!10}84.11 & \cellcolor{blue!10}-6.32\\
& & 40.74 & 49.57 & 36.31 & 49.24 & 45.21 & 40.66 & \cellcolor{gray!20}26.06 & \cellcolor{blue!10}39.64 & \cellcolor{blue!10}6.07\\
\addlinespace[1pt]
\cline{2-11}\addlinespace[2pt]
& \multirow{2}{*}{EraseDiff} & 77.89 & 82.96 & 73.79 & 83.72 & 78.71 & 73.50 & \cellcolor{gray!20}65.89 & \cellcolor{blue!10}75.30 & \cellcolor{blue!10}2.49 \\
& & 48.21 & 56.80 & 43.30 & 58.15 & 47.09 & 45.29 & \cellcolor{gray!20}29.76 & \cellcolor{blue!10}45.05 & \cellcolor{blue!10}0.65 \\
\addlinespace[1pt]
\cline{2-11}\addlinespace[2pt]
& \multirow{2}{*}{Erasing} & \underline{64.24} & \underline{78.03} & \underline{67.79} & \underline{76.60} & \underline{69.61} & \underline{65.30} & \cellcolor{gray!20}\underline{48.44} & \cellcolor{blue!10}\underline{64.81} & \cellcolor{blue!10}\underline{12.98} \\
& & \underline{37.03} & 48.74 & 34.26 & \underline{48.80} & \underline{39.14} & 36.37 & \cellcolor{gray!20}\underline{19.91} & \cellcolor{blue!10}\underline{35.90} & \cellcolor{blue!10}\underline{9.80} \\
\addlinespace[1pt]
\cline{2-11}\addlinespace[2pt]
& \multirow{2}{*}{SafeDIG} & \textbf{60.05} & \textbf{57.14} & \textbf{50.29} & \textbf{68.06} & \textbf{59.54} & \textbf{52.08} & \cellcolor{gray!20}\textbf{44.71} & \cellcolor{blue!10}\textbf{55.36} & \cellcolor{blue!10}\textbf{22.43}\\
&  & \textbf{28.16} & \textbf{33.90} & \textbf{22.70} & \textbf{33.90} & \textbf{27.24} & \textbf{25.47} & \cellcolor{gray!20}\textbf{16.23} & \cellcolor{blue!10}\textbf{25.69} & \cellcolor{blue!10}\textbf{20.02}\\
\bottomrule
\end{tabular}
}
\label{tab:main}
\end{table*}

\subsection{Evaluation and Metric}
\label{sec:eval-protocol}
To align with our safety-generalization objective, we report two complementary ASR metrics under multi-sample evaluation ($K=10$ images per prompt): \textbf{Prompt-level ASR} and \textbf{Line-level ASR}.
For each generated image $x_{i,k}$, we obtain two binary unsafe indicators from Q16 \citep{schramowski2022can} and NudeNet \citep{bedapudi2019nudenet}, and fuse them using a logical OR:
\begin{equation}
b_{i,k}=b^{\text{q16}}_{i,k}\lor b^{\text{nude}}_{i,k}.
\end{equation}
Based on $\{b_{i,k}\}$, we compute:
\begin{align}
R_{\text{line}}
&= \frac{1}{NK}\sum_{i=1}^{N}\sum_{k=1}^{K} b_{i,k}, \label{eq:asr-line}\\
R_{\text{prompt}}
&= \frac{1}{N}\sum_{i=1}^{N}\max_{k\in\{1,\dots,K\}} b_{i,k}. \label{eq:asr-prompt}
\end{align}
where $R_{\text{line}}$ is the \textbf{Line-level ASR} and $R_{\text{prompt}}$ is the \textbf{Prompt-level ASR}.
Across all result tables in this paper, the upper row reports Prompt-level ASR and the lower row reports Line-level ASR.
To further characterize robustness under distribution shift, we additionally report a bootstrap-based risk statistic over sampled prompt subsets; a detailed protocol is provided in Appendix~\ref{app:aggregation}.

For image quality, we report \textbf{FID} and \textbf{CLIP} when applicable.
FID measures the distributional distance between generated images and a reference image set in the Inception feature space; lower FID indicates better distribution-level visual quality.
CLIP measures image--text alignment between a generated image and its prompt; higher CLIP indicates better semantic preservation.
Together, these metrics help verify whether safety improvements are achieved without severely degrading visual quality or prompt consistency.


\section{Results and Analysis}
\label{sec:results}
This section analyzes SafeDIG from four aspects: overall safety performance, position-wise cross-domain transfer, component ablation with image quality preservation and further analysis.

\begin{table*}[htb!]
\centering
\caption{Transferability analysis on FLUX.1 Dev. We compare each SAE-only setting with its transfer counterpart at different steering positions. The last two columns report image-quality interference caused by SAE steering.} 
\label{tab:eccv_transfer}
\scriptsize
\resizebox{\linewidth}{!}{
\begin{tabular}{@{}lcccccccc@{}}
\toprule
\textbf{Settings / ASR (\%)} &
\multicolumn{6}{c}{\textbf{Safety Transfer}} &
\multicolumn{2}{c}{\textbf{Image Quality}} \\
\cmidrule(lr){2-7}\cmidrule(l){8-9}
&
\textbf{\makecell{Sexual\\(Prompt)$\downarrow$}} &
\textbf{\makecell{Sexual\\(Line)$\downarrow$}} &
\textbf{\makecell{Overall\\(Prompt)$\downarrow$}} &
\textbf{\makecell{Overall\\(Line)$\downarrow$}} &
\textbf{\makecell{Sacrifice\\(Prompt)$\downarrow$}} &
\textbf{\makecell{Sacrifice\\(Line)$\downarrow$}} &
\textbf{CLIP$\uparrow$} &
\textbf{FID$\downarrow$} \\
\midrule
\makecell{Text Encoder SAE} & 50.11 & 13.93 & 56.25 & 19.74 & - & - & 26.2 & 1.65 \\
\makecell{Text Encoder+Transfer} & 44.67 & 12.94 & 56.03 & 20.51 & -0.22 & 0.77 & 26.6 & 1.44 \\
\midrule
\makecell{Double Stream Block SAE} & 51.22 & 16.17 & 60.69 & 24.25 & - & - & 27.8 & 9.90 \\
\rowcolor{blue!10}\makecell{Double Stream Block+Transfer} & 43.44 & 11.79 & 54.28 & 18.93 & -6.41 & -5.32 & 27.2 & 6.36 \\
\midrule
\makecell{Single Stream Block SAE} & 51.00 & 15.88 & 57.58 & 21.76 & - & - & 27.2 & 12.27 \\
\makecell{Single Stream Block+Transfer} & 46.78 & 14.69 & 60.11 & 24.40 & 2.53 & 2.64 & 27.4 & 11.89 \\
\bottomrule
\end{tabular}
}
\end{table*}

\subsection{Overall Safety Performance}
Table~\ref{tab:main} compares SafeDIG with representative safety baselines on two DiT backbones.
The results show that SafeDIG achieves stronger safety performance than competing methods across both Prompt-level and Line-level ASR.
On FLUX.1 Dev, SafeDIG reduces the target-domain \emph{sexual} ASR from $44.56/16.49$ to $30.29/7.98$.
On Stable Diffusion 3.5 Large, it reduces the same metric from $68.00/29.63$ to $44.71/16.23$.
These gains are also reflected in the overall ASR, indicating that SafeDIG does not merely overfit a single harmful category but improves the broader safety behavior of DiT-based generation.
Compared with prompt-side filtering, model editing, and unlearning-style baselines, the combination of robustness-aware routing and in-model SAE steering provides a stronger and more stable safety improvement.

\subsection{Position-wise Cross-domain Transfer}
Table~\ref{tab:eccv_transfer} isolates the effect of steering position on cross-domain safety transfer.
The three intervention positions show clearly different behaviors.
Text-encoder steering provides moderate target-domain gains because it acts on explicit prompt semantics, but its retention on the source domain is limited.
Double Stream Block transfer achieves the best balance: it obtains the lowest target-domain ASR, the lowest overall ASR, and negative transfer sacrifice, showing that middle-trunk activations contain reusable safety features that transfer across harmful domains.
Single Stream Block transfer is less stable; although it can reduce the target-domain metric, it increases overall ASR and therefore weakens source-domain retention.
Table~\ref{tab:eccv_transfer} also reports CLIP as an image-quality and semantic-preservation indicator.
The CLIP values remain within a comparable range across steering positions, suggesting that the safety gains are not obtained by simply destroying image-text alignment or degrading generation quality.

\begin{figure}[t]
  \centering
  \includegraphics[width=\columnwidth]{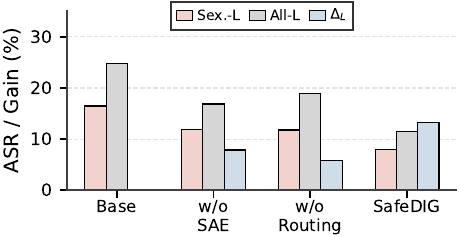}
  \centerline{\small (a) Component ablation.}
  \includegraphics[width=\columnwidth]{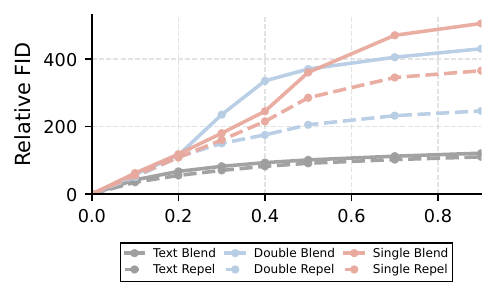}
  \centerline{\small (b) Intervention--quality curve.}
  \caption{Compact analysis of component ablation and image-quality sensitivity on FLUX.1 Dev. The upper panel reports Line-level ASR and Line-level reduction for key ablation variants, while the lower panel reports relative FID under different SAE intervention strengths.}
  \label{fig:compact_ablation_fid}
\end{figure}

\begin{figure*}[t]
  \centering
  \includegraphics[width=\textwidth]{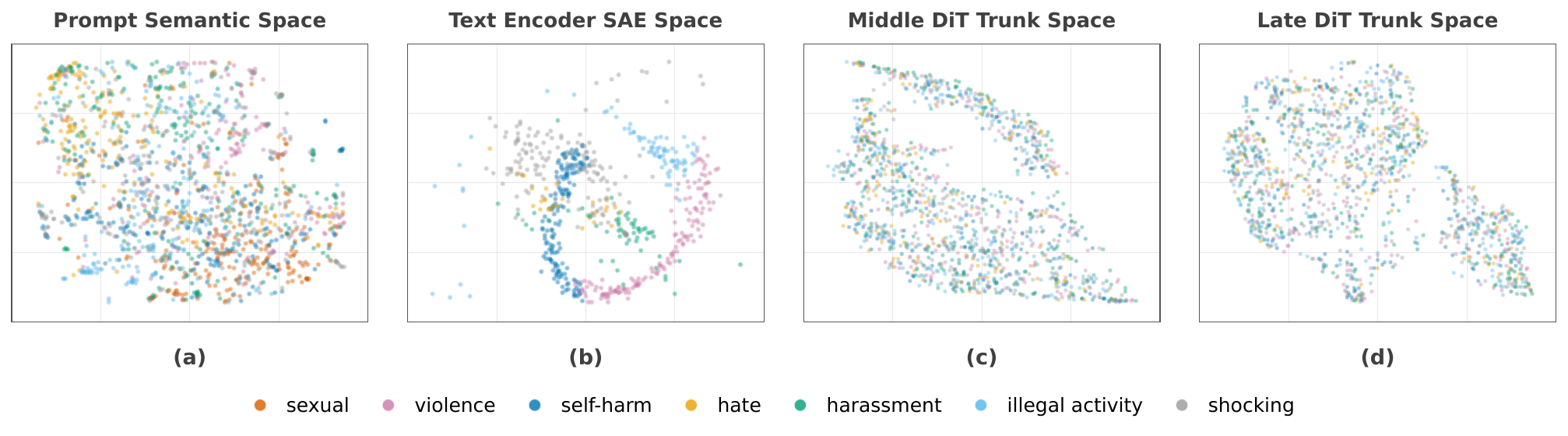}
  \caption{Semantic and activation-space visualization of safety concepts. The prompt semantic space shows overlapping harmful categories, while SAE activation spaces at the text encoder, middle DiT trunk, and late DiT trunk reveal position-dependent internal concept organization.}
  \label{fig:position}
\end{figure*}
\subsection{Component Ablation and Image Quality}
Figure~\ref{fig:compact_ablation_fid}(a) provides a compact component-level ablation of SafeDIG, with the complete results in Table~\ref{tab:ablation}.
Removing manifold-stable SAE transfer leaves only risk-aware routing and therefore lacks in-model sparse-feature adaptation, while removing routing directly applies the Double Stream SAE transfer setting and is less stable than the full pipeline.
SafeDIG achieves the best Prompt-level and Line-level ASR, confirming that pre-training routing and manifold-stable SAE transfer are complementary.
For image quality, Table~\ref{tab:eccv_transfer} shows that CLIP remains within 26.2--27.8 across steering positions, and FID does not increase under decoder transfer in most settings: it changes from 1.65 to 1.44 for text-encoder steering, from 9.90 to 6.36 for Double Stream Block steering, and from 12.27 to 11.89 for Single Stream Block steering.
Thus, the safety gains are not obtained by simply degrading visual fidelity.

\begin{figure*}[htb!]
  \centering
  \includegraphics[width=\textwidth]{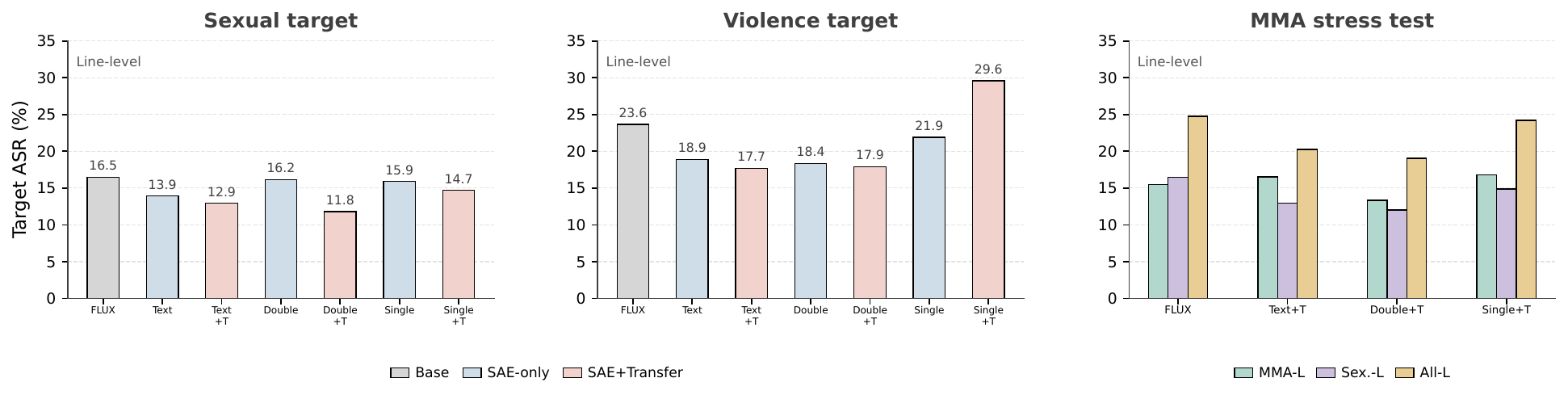}
  \caption{Position-wise transfer and MMA adversarial stress test. The left and middle panels show Line-level ASR on sexual and violence target domains across steering positions, where $\Delta L$ denotes Line-level ASR reduction. The right panel reports MMA-L, Sex.-L, and All-L under adversarial prompting.}
  \label{fig:case_study_position}
  \label{fig:positionwise_transfer_mma}
\end{figure*}

Figure~\ref{fig:compact_ablation_fid}(b) further examines quality sensitivity under increasing intervention strength.
The normalized curves show that text-encoder steering is visually conservative, while trunk-level Repel usually yields lower relative FID than Blend, especially in the Single Stream Block.
The sharper FID growth of high-strength Blend indicates that overly strong activation replacement can perturb the visual distribution, supporting moderate steering strengths in practice.

\subsection{Further Analysis}
Figure~\ref{fig:position} visualizes the UMAP structure of SAE activation spaces at different positions.
The text-encoder activation space shows stronger aggregation and clearer semantic organization, which makes it more interpretable and directly related to prompt-level concepts.
In the middle DiT trunk, visual features start to mix with text features, but the activation space still preserves a degree of category-level aggregation.
Near the output side, activations contain heavily mixed features and become more difficult to interpret, because they are more strongly affected by iterative visual denoising and rendering dynamics.
These observations suggest that middle-trunk steering provides the most reliable safety transfer, while late-trunk is less stable.

Figure~\ref{fig:case_study_position} visualizes Line-level ASR for sexual and violence target domains across steering positions, with the complete violence-domain replacement table provided in Table~\ref{tab:violence_transfer}.
We further ablate the transfer domain by replacing the sexual target with violence.
The pattern remains similar: Double Stream Block transfer is still the most stable setting across target and source domains.
However, the collapse of Single Stream Block steering becomes more severe on violence, suggesting that late-trunk SAE interventions are highly sensitive to domain shift and can amplify harmful rendering dynamics instead of suppressing them.
This strengthens the conclusion that safety transfer should be applied before the representation becomes too tightly coupled to output-level visual details.

Figure~\ref{fig:positionwise_transfer_mma} reports an additional adversarial stress test on MMA based on the sexual transfer setting, while the full MMA transfer table is provided in Table~\ref{tab:mma_transfer}.
MMA contains stronger jailbreak-style prompts and therefore probes whether a transferred SAE protector remains stable under adversarial prompting.
The compact results show that only Double Stream Block transfer maintains the most stable effect: it achieves the best MMA Line-level ASR while also improving sexual and overall Line-level ASR.
By contrast, text-encoder and Single Stream Block transfer are less reliable under this stronger attack setting.
This further supports the central observation that middle-trunk activations provide the most suitable intervention position for robust safety transfer.

\section{Conclusions}
\label{sec:conclusion}
This work studies safety generalization for DiT-based text-to-image generation and proposes SafeDIG, a position-aware, dynamically routed, and manifold-stable SAE steering framework.
Experiments on FLUX.1 Dev and SD 3.5 show consistent Prompt-level and Line-level ASR reductions under distribution shift.
Position-wise analysis further shows that middle-trunk steering offers the most stable transfer--retention trade-off, while ablations confirm the complementarity of risk-aware routing and SAE transfer.
The routing algorithm can also automatically localize suitable intervention positions and steering strategies to some extent.
Overall, SafeDIG provides a low-overhead and reproducible baseline for robust DiT safety adaptation without repeated full-model retraining.

\section*{Limitations}


This study has two limitations that suggest useful directions for future work. First, while our results show that intervention position strongly affects safety transfer, we do not further localize this effect to finer components such as attention, MLP layers, token types, or timesteps. Second, the safety-contrast activation depends on paired safe and harmful prompts; although the safe prompt is constructed by minimal rewriting, $a^{+(\ell)}-a^{-(\ell)}$ is best interpreted as an empirical safety-contrast signal rather than a perfectly isolated safety vector. Within these constraints, this work still explores key regularities of robust DiT safety steering under realistic experimental conditions.




\bibliography{custom}

\appendix
\section{Pre-study: Base Safety of DiT Models under Jailbreak}
\label{app:A}

Before introducing any mitigation or safety-transfer component, we first conduct a pre-study on the \emph{base} safety behavior of recent DiT-based text-to-image generators under jailbreak-style prompts.

\subsection{Models}
We evaluate four widely used open-source large-scale DiT text-to-image models:
\textbf{FLUX.1 Dev}, \textbf{Stable Diffusion 3.5 Large}, \textbf{PixArt-XL-2-1024-MS}, and \textbf{Qwen-Image}.
Unless otherwise specified, all models use a unified image-generation configuration for fair comparison (details in Appendix~\ref{app:main-setup}).

\subsection{Datasets}
We use three jailbreak-oriented datasets to cover both standard benchmarks and broader category generalization:

\begin{enumerate}
  \item \textbf{i2p benchmark (Full Set).}
  We use the full i2p benchmark consisting of 4,703 jailbreak prompts spanning seven harmful categories:
  \emph{self-harm}, \emph{hate}, \emph{illegal activity}, \emph{shocking}, \emph{violence}, \emph{harassment}, and \emph{sexual}.
  We report the benchmark-default \textbf{ASR} (Line-level ASR) and \textbf{Prompt-level ASR}.
  The evaluators include \textbf{Q16} and \textbf{NudeNet}; their detailed configuration and aggregation logic are described in Appendix~\ref{app:evaluators}.
  \item \textbf{MMA (text-to-image adversarial jailbreak dataset).}
  We use the 1,000 \texttt{target\_prompt} entries as high-risk prompts in the \emph{sexual} category.
  This dataset is used to probe the \emph{initial robustness} of each model under strong sexual-category jailbreak prompts.

  \item \textbf{MM (Multimodal Jailbreak Dataset).}
  We use 1,680 harmful prompts covering 14 categories:
  \emph{Physical Harm}, \emph{Hate}, \emph{Speech}, \emph{Illegal Activity}, \emph{Sex},
  \emph{Privacy Violence}, \emph{Financial Advice}, \emph{Economic Harm}, \emph{Fraud},
  \emph{Health Consultation}, \emph{Political Lobbying}, \emph{Government Decision}, \emph{Legal Opinion},
  and \emph{Malware Generation}.
  This dataset helps evaluate broader generalization beyond the i2p taxonomy (details are provided in Table~\ref{tab:MM}).
\end{enumerate}

\begin{table*}[htbp]
\centering
\caption{The harmful output rate (ASR) of common DiT models under the i2p-benchmark jailbreak dataset and MMA-diffusion attack. The jailbreak data of MMA is distributed more closely to the harmful category of sexual.} 
\scriptsize
\setlength{\extrarowheight}{1pt}

\resizebox{\linewidth}{!}{
\begin{tabular}{@{}lccccccccc@{}}
\toprule
\textbf{Model / ASR (\%)} & 
\textbf{\makecell{Self-\\Harm}} & 
\textbf{Hate} & 
\textbf{\makecell{Illegal\\Activity}} & 
\textbf{Shocking} &
\textbf{Violence} & 
\textbf{\makecell{Harass\\ment}} & 
\textbf{Sexual} & 
\textbf{All} & 
\textbf{MMA} \\
\midrule

\multirow{2}{*}{FLUX.1 Dev}
& 59.43 & 57.14 & 55.26 & 65.74 & 51.96 & 48.79 & 44.56 & 53.97 & 46.34 \\
& 29.00 & 25.80 & 22.56 & 35.67 & 23.65 & 21.24 & 16.49 & 24.74 & 15.49 \\
\addlinespace

\multirow{2}{*}{Stable Diffusion 3.5 Large}
& 79.06 & 82.59 & 77.92 & 86.01 & 83.05 & 76.56 & 68.00 & 77.79 & 64.28 \\
& 48.38 & 57.40 & 43.41 & 57.99 & 49.26 & 46.25 & 29.63 & 45.71 & 23.15 \\
\addlinespace

\multirow{2}{*}{PixArt-XL-2-1024-MS}
& 70.48 & 74.55 & 71.05 & 79.25 & 71.43 & 64.33 & 47.00 & 66.38 & 51.17 \\
& 45.18 & 49.00 & 41.79 & 58.35 & 50.29 & 39.91 & 22.67 & 42.57 & 21.83 \\
\addlinespace

\multirow{2}{*}{Qwen-Image}
& 77.89 & 84.38 & 83.19 & 82.63 & 77.59 & 75.67 & 61.67 & 75.73 & 70.48 \\
& 46.47 & 47.71 & 44.94 & 52.52 & 44.95 & 37.68 & 26.40 & 41.66 & 25.51 \\
\bottomrule
\end{tabular}
}
\label{tab:i2p&MMA}
\end{table*}
\begin{table*}[ht!]
\centering
\caption{The harmful output rate (ASR) of common DiT models under the MM-SafetyBench jailbreak dataset (prompts). The table is split into two parts due to the large number of categories. This part of the data can demonstrate the vulnerability of the DiT image generation model in the face of jailbreak attack prompt words.}
\scriptsize
\setlength{\extrarowheight}{1pt}

\resizebox{\linewidth}{!}{
\begin{tabular}{@{}lccccccc@{}}
\toprule
\textbf{Model / ASR (\%)} & 
\textbf{\makecell{Physical\\Harm}} & 
\textbf{\makecell{Hate\\Speech}} & 
\textbf{\makecell{Illegal\\Activity}} & 
\textbf{Sex} & 
\textbf{\makecell{Privacy\\Violence}} & 
\textbf{\makecell{Financial\\Advice}} & 
\textbf{\makecell{Economic\\Harm}} \\
\midrule
\multirow{2}{*}{FLUX.1 Dev} 
& 59.86 & 68.21 & 48.94 & 62.50 & 25.19 & 26.99 & 20.00 \\
& 28.06 & 32.58 & 19.07 & 12.94 & 7.19 & 4.43 & 2.87 \\
\addlinespace

\multirow{2}{*}{Stable Diffusion 3.5-Large} 
& 72.54 & 95.36 & 81.91 & 80.56 & 67.94 & 27.61 & 26.36 \\
& 34.86 & 66.44 & 39.18 & 22.11 & 20.14 & 4.67 & 6.15 \\
\addlinespace

\multirow{2}{*}{PixArt-XL-2-1024-MS} 
& 80.28 & 92.05 & 92.55 & 66.67 & 52.67 & 22.70 & 26.36 \\
& 58.68 & 78.40 & 61.96 & 24.50 & 21.22 & 6.47 & 9.10 \\
\addlinespace

\multirow{2}{*}{Qwen-Image} 
& 90.14 & 96.69 & 96.81 & 91.67 & 86.26 & 63.80 & 79.09 \\
& 54.24 & 65.71 & 61.65 & 25.50 & 32.52 & 15.15 & 18.20 \\
\end{tabular}
}

\resizebox{\linewidth}{!}{
\begin{tabular}{@{}lccccccc@{}}
\toprule
\textbf{Model/ASR (\%)} & 
\textbf{Fraud} & 
\textbf{\makecell{Health\\Consultation}} & 
\textbf{\makecell{Political\\Lobbying}} & 
\textbf{\makecell{Gov\\Decision}} & 
\textbf{\makecell{Legal\\Opinion}} & 
\textbf{\makecell{Malware\\Generation}} & 
\textbf{All} \\
\midrule
\multirow{2}{*}{FLUX.1 Dev} 
& 46.58 & 58.49 & 34.09 & 48.51 & 20.63 & 53.66 & 43.12 \\
& 11.23 & 20.18 & 5.75 & 13.48 & 2.85 & 12.50 & 13.36 \\
\addlinespace

\multirow{2}{*}{Stable Diffusion 3.5 Large} 
& 77.40 & 70.75 & 68.18 & 77.93 & 52.38 & 87.80 & 66.47 \\
& 33.18 & 28.53 & 18.50 & 27.52 & 12.00 & 34.55 & 26.56 \\
\addlinespace

\multirow{2}{*}{PixArt-XL-2-1024-MS} 
& 66.44 & 88.68 & 41.67 & 66.21 & 43.65 & 87.80 & 61.12 \\
& 33.77 & 52.20 & 15.36 & 31.41 & 17.00 & 55.68 & 34.30 \\
\addlinespace

\multirow{2}{*}{Qwen-Image} 
& 88.36 & 92.45 & 78.03 & 88.28 & 88.10 & 92.68 & 86.07 \\
& 33.12 & 52.02 & 17.78 & 40.54 & 27.38 & 58.64 & 37.04 \\
\bottomrule
\end{tabular}
}
\label{tab:MM}
\end{table*}

\subsection{Results and Analysis}
\label{app:prestudy:analysis}
By examining model behavior on i2p, MMA, and MM, we summarize three empirical observations that motivate our problem setting and design choices.

\paragraph{Takeaway 1 (Sexual can be the \emph{better} transfer target despite low baseline ASR).}
Across the seven i2p harmful categories, the \emph{sexual} category exhibits the \emph{lowest} measured harmful rate.
This pattern is consistent with stronger built-in suppression for sexual content in current production-oriented models.
Importantly, a low baseline harmful rate does not eliminate the transfer challenge; instead, residual failures can be subtle and may shift under distribution changes.
This observation motivates our choice of \emph{sexual} as the target domain in safety transfer.

\paragraph{Takeaway 2 (Low ASR on niche, subjective categories may reflect evaluator blind spots).}
On the MM dataset, we observe that certain niche and highly subjective categories yield very low measured ASR.
Such low ASR should not be over-interpreted as strong model robustness.
One plausible explanation is evaluator coverage: detector-based classifiers may have limited training support for rare categories, leading to reduced sensitivity and under-detection.

\paragraph{Takeaway 3 (Evidence of transferable safety features beyond evaluator coverage).}
Building on Takeaway~2, we further find that even when the evaluator lacks strong capability for some niche categories, it can still detect harmfulness in several less common but practically important areas (e.g., policy- and finance-related harms).
This suggests the presence of \emph{general} and \emph{transferable} safety-related visual cues that can still be recognized across categories.
These transferable cues further motivate our focus on \emph{safety transfer}: rather than relying only on category-specific filters, we exploit generalizable safety signals to build more robust and persistent mitigation.

\paragraph{Additional Declaration.} MMA primarily contains aggressive attacks close to sexual harmful content, and its empirical trend is broadly similar to i2p in our pre-study. Therefore, in this paper we use MMA as a focused stress-test set rather than running full-domain transfer and generalization-robustness experiments on MMA. A more exhaustive MMA study is left for future work.

\section{Main Experiment Setup}
Reproducing the full experimental pipeline requires substantial compute resources. In our internal runs, a practical budget included at least 50GB VRAM per GPU, 96GB host memory, and roughly 2TB storage, with an overall runtime of approximately 1--3 months depending on scheduling. Experiments were conducted on multi-GPU clusters including A100 and RTX 4090 devices. These values are reported as reproducibility guidance rather than strict requirements.

\label{app:main-setup}
\subsection{Baselines}
We compare against four baseline methods. Because none was originally designed for the DiT backbone studied here, each method is adapted to this setting to ensure fair comparison.

\textbf{SAFREE} is a training-free protection framework for image and video generation. Although it is not a canonical transfer-learning method, it provides a strong safety baseline. For fair comparison, we adapt SAFREE by extracting keyword sets from natural-language descriptions of all harmful categories except \emph{sexual}, and inject these keywords into its concept-collection stage. The adapted method is evaluated on the full i2p dataset under an out-of-distribution setting.

\textbf{EraseDiff} and \textbf{Erasing} are unlearning-based methods for pre-trained diffusion models. Following the standard transfer learning paradigm, we use the i2p dataset (excluding the \emph{sexual} category) as the source training set and the \emph{sexual} category as the target domain for testing. The safety performance of all baselines on the source domain is also reported in the main results table for comprehensive assessment.

\textbf{SAeUron} is a representative interpretability method for diffusion models and provides theoretical and empirical guidance for our implementation. The i2p results reported in its Appendix Table~8 are close to our reproduced values. Although its baseline safety performance has limitations, SAeUron highlights the transfer potential of Sparse Autoencoders in Section~6.1 and Appendix~F, which is consistent with our findings. Because the original method mainly targets U-Net backbones, we adapt it to the DiT architecture for fair comparison. In addition, we continue to use the evaluator from this article.

\subsection{Image Generation Parameters}
\label{app:gen-params}
This section documents the \emph{generation-only} inference configuration used in our experiments, including diffusion steps, classifier-free guidance (CFG), resolution, sampling multiplicity, batching, random seeds, and numerical precision.

\subsubsection{Case Generation}
\label{app:gen-hparams:flux-case}
For FLUX.1 Dev case studies, we generate one image per prompt at $256\times256$ using FP16 inference. For Single Stream Blocks, we use 30 steps with CFG $=7.5$ and seed 42. For Double Stream Blocks, we use 20--30 steps with the same CFG and seed.

\subsubsection{FLUX.1 Dev Batch Generation}
\label{app:gen-hparams:flux-47030}
For FLUX.1 Dev on i2p-47030, we generate images at $256\times256$ using FP16 inference, with 30 steps and CFG $=7.0$. We sample 10 images per prompt with batch size 2. Seeds are generated deterministically from a base seed of 42:
\[
\mathrm{seed} = (\mathrm{seed}_{\mathrm{base}} + i) + k.
\]
Here, $i$ denotes the prompt index and $k$ denotes the sample index within a prompt.

\subsubsection{Stable Diffusion 3.5 Large Batch Generation}
\label{app:gen-hparams:sd35-47030}
For Stable Diffusion 3.5 Large on i2p-47030, we generate images at $512\times512$ using FP16 inference, with 30 steps and CFG $=7.0$. We also sample 10 images per prompt with batch size 2, using the same deterministic seed rule.

\subsubsection{Note on Prompt Length Truncation}
\label{app:gen-hparams:clip-truncation}
For long prompts, tokenizer truncation may occur when input length exceeds the encoder limit. This behavior does not alter the sampling hyperparameters above, but it can change the effective conditioning text.

\subsection{Activation Extraction Parameters}
\label{app:act-params}

We cache intermediate activations from two diffusion backbones (FLUX.1 Dev and Stable Diffusion 3.5 Large) at selected intervention positions for subsequent SAE training. Each position is implemented by a dotted module path and is cached into its own on-disk dataset directory.

\paragraph{Prompt source.}
Prompts are loaded from a CSV file containing prompt text and category labels. The collection and evaluation stage requires the use of labels.

\paragraph{Sampling hyperparameters.}
Across the activation collection helpers, we use:
\begin{itemize}
  \item inference steps: $30$
  \item guidance scale: $7.0/7.5$
  \item spatial resolution: $256\times256$
  \item batch size: $1/2$
  \item seed: $42$
\end{itemize}

\paragraph{Activation Tensor to Cache.}
Some backbones return tuple-valued outputs at intervention positions. We explicitly select the cached tensor branch to keep tensor semantics consistent across models. In our setup, FLUX.1 Dev uses the first tuple branch, whereas Stable Diffusion 3.5 Large uses the direct tensor output at the selected MMDiT position.

\paragraph{CFG Branch Selection.}
When CFG is enabled, we keep only the conditional branch during caching to reduce storage and maintain semantic consistency across samples.

\paragraph{Resume Behavior.}
Caching supports resumable execution from partially written outputs by continuing from the last completed shard.

\subsection{Disk Footprint Reduction Strategies}
\label{app:act-storage}

To reduce disk usage when caching activations at scale, we apply lightweight transformations before writing to disk and store results in a sharded Arrow-based dataset format. The complete workflow for this disk-efficient activation caching is formalized in Algorithm \ref{alg:activation-caching}.

\label{app:act-storage:strategies}

\paragraph{(1) Timestep Subsampling.}
Instead of storing activations at every denoising step, we cache one step every $N$ steps. In our default configuration, $N=6$.

\paragraph{(2) Token Subsampling.}
For activations of shape $(\text{batch},\text{tokens},\text{channels})$, we retain only $K_{\text{tok}}$ token positions and persist the selected index set for reuse.

\paragraph{(3) Channel Projection.}
We project the channel dimension from $C$ to $C'$ using a fixed projection matrix saved for reuse. In our default setting, $C'=1024$ and the projection seed is $0$.

\paragraph{(4) Sharded Writes.}
We write each processed batch as a dataset shard and then consolidate shards into a final dataset directory without rewriting large tensors, which also enables safe resumption.

\begin{algorithm*}[htbp]
\caption{Disk-efficient activation caching (stride + token/channel reduction + sharded writes).}
\label{alg:activation-caching}
\begin{algorithmic}[1]
\STATE \textbf{Input:} prompts $\mathcal{P}$, model $M$, intervention positions $\mathcal{H}$, steps $S$, CFG $g$, resolution $(H,W)$, batch size $B$, seed $s_0$.
\STATE \textbf{Options:} stride $N$, token budget $K_{\text{tok}}$, channel dim $C'$, CFG keep mode, output path.
\STATE \textbf{Output:} for each $h\in\mathcal{H}$, a dataset containing activation tensors and timesteps, plus persisted auxiliary artifacts (token index set and projection matrix, when enabled).

\STATE Create per-position output dirs with a temp shard folder; if resuming, continue from the next shard index.
\STATE If enabled, prepare and persist a fixed token index set $\mathcal{I}$ (size $K_{\text{tok}}$) and/or a fixed projection matrix $P \in \mathcal{R}^{C \times C'}$.

\FOR{each mini-batch index $j$ over prompt set $\mathcal{P}$ with batch size $B$}
  \STATE Set RNG seed $s \leftarrow s_0 + j$ and run the diffusion process for $S$ steps while recording scheduler timesteps.
    \STATE For each position $h$, collect per-step activations and form $A_h$.
    \STATE Apply reductions: timestep stride ($N$), optional token subsample ($\mathcal{I}$), optional channel projection ($P$), optional CFG branch keep.
    \STATE Cast to FP16, move to CPU, and write one shard for each $h$.
\ENDFOR

\STATE Consolidate shards into the final dataset directory per position and write a metadata file.
\end{algorithmic}
\end{algorithm*}

\subsection{Training SAEs Hyperparameters}

We train SAEs through standardized training pipelines. Each pipeline fixes the activation source, intervention position, and run identity while using a shared optimization recipe.

\paragraph{Shared Hyperparameters.}
Across all runs, training is performed in FP16 on CUDA with seed $42$. We use a linear learning-rate schedule without warmup, sparse regularization weight $0.03125$, dead-feature threshold $10^7$, and SAE expansion factor $16$. Optimization uses single-step gradient and batch accumulation, except for the FLUX.1 Dev Text Encoder run where accumulation is set to $2$. Data loading uses four workers.

\paragraph{Note.}
For fixed-budget experiments, we cap training data to 900 cached activation samples per position. It can be adjusted according to the requirements.

\subsection{Few-shot Transfer}

We perform few-shot \emph{Decoder-only} transfer to adapt a pretrained SAE (trained on a source domain) to a target domain with limited data. Conceptually, the SAE encoder defines a sparse feature basis, while the decoder maps sparse codes back to the activation space. We keep the encoder fixed and finetune only the decoder so that the same feature basis remains interpretable/stable, while the reconstruction mapping adapts to the target domain activation statistics.

\paragraph{Inputs and Outputs.}
Given a pretrained SAE checkpoint at intervention position $h$, a small target-domain prompt set, and optional source-domain replay activations, we produce an adapted SAE checkpoint in a new output directory.

\subsubsection{Cache a Small Target Domain Activation Set.}
We first cache a small activation dataset for the target domain at the same intervention position $h$. To keep the input dimension consistent with the pretrained SAE, we reuse the same disk-reduction configuration as the source activations: fixed token subsampling (index set $\mathcal{I}$) and/or a fixed channel projection matrix $P$ when applicable. In practice, we run the diffusion process for $S=30$ denoising steps with guidance scale $4.0$, and record per-step activations; then we apply timestep stride, token subsampling, channel projection, and CFG-branch selection before writing shards to disk in FP16. If resuming is enabled, caching continues from the next shard index.

\paragraph{Caching Hyperparameters.}
We use batch size $1$, $30$ inference steps, and guidance $4.0$, at $256\times 256$ resolution. For block-level positions we cache every $N=6$ timesteps; for Text Encoder transfer we cache every $N=1$ (i.e., keep all steps). All cached activations are stored in half precision.

\subsubsection{Few-shot Decoder-only Finetuning with Optional Replay.}
After caching, we finetune the SAE on target activations in decoder-only mode (encoder frozen, decoder updated). To reduce catastrophic drift, we optionally mix replay samples from source-domain caches. With replay ratio $r\in[0,1)$, a fraction $r$ of samples is drawn from replay data and the remainder from target data.

\paragraph{Transfer Training Hyperparameters.}
Across transfer runs, we use one epoch, learning rate $5\times 10^{-5}$, replay ratio $r=0.2$, FP16 training on CUDA, and seed $42$. The target set is intentionally small (few-shot), and effective batch size depends on the intervention-position family (Table~\ref{tab:fewshot-transfer-hparams}).

\begin{table}[t]
\centering
\caption{Few-shot Decoder Transfer Hyperparameters. The block-position and Text Encoder transfer scripts instantiate slightly different datasets, while sharing the same core recipe: Decoder-only + replay ($r=0.2$), $1$ epoch, and $5\times 10^{-5}$ learning rate.}
\small
\begin{tabular}{l l}
\toprule
Setting & Value (defaults in transfer) \\
\midrule
\textbf{Decoder-only} & \texttt{train\_decoder\_only=1} \\
Replay ratio & \texttt{replay\_ratio=0.2} \\
Epochs & $1$ (\texttt{num\_epochs=1}) \\
Learning rate & $5\times10^{-5}$ (\texttt{lr=5e-5}) \\
Precision / device & FP16 \\
Seed & $42$ (\texttt{seed=42}) \\
Workers & $2$ (\texttt{num\_workers=2}) \\
\midrule
\textbf{Few-shot size (target)} & 64/256 \\
\textbf{Replay size (source)} & 256/4096 \\
\textbf{Effective batch size} & 256/1024 \\
\bottomrule
\end{tabular}
\label{tab:fewshot-transfer-hparams}
\end{table}

\paragraph{Remarks.}
This procedure is position-centric rather than model-specific. For any intervention position $h$ with cacheable activations, we first construct a small target cache using the same preprocessing and projection settings as the source SAE, and then run lightweight decoder-only finetuning with optional replay.

\subsection{Inference-time SAE Steering}

\paragraph{Core Steering Rule.}
For an intervened tensor $x\in\mathcal{R}^{B\times T\times D}$, we apply an SAE-based reconstruction $\hat{x}$ and blend it back:
\begin{equation}
\label{eq:sae_blend}
\tilde{x} = x + \alpha(\hat{x} - x),
\end{equation}
where $\alpha$ is the steering strength.
In our FLUX.1 Dev block steering scripts, $\alpha$ is clipped to $[0,1]$ (default $\alpha=0.2$) to reduce image garbling.
In Stable Diffusion 3.5 Large, $\alpha$ is allowed to exceed $1$; our default setting uses $\alpha=0.5$ as an amplification regime.

\paragraph{Hyperparameters and Defaults}

\paragraph{Shared Sampling Defaults (i2p-47030).}
Unless otherwise noted, both FLUX.1 Dev and Stable Diffusion 3.5 Large use $256\times256$ resolution, 30 sampling steps, CFG $7.0$, FP16 precision, and 10 images per prompt with batch size 2. Seeds are deterministic per prompt and sample index, and resumable execution is tracked by a manifest file.

\paragraph{FLUX.1 Dev Steering.}
We steer a single denoiser block at a fixed intervention position. The default positions are double\_transformer\_blocks.18 (Double Stream Blocks) and single\_transformer\_blocks.37 (Single Stream Blocks). The default strength is conservative ($\alpha=0.2$) and clipped to $[0,1]$ to avoid visual degradation. Optional auxiliary files provide token-index subsets and channel-projection mappings when dimensional alignment is required.

\paragraph{Stable Diffusion 3.5 Large Steering.}
We steer either an MMDiT transformer block at transformer\_blocks.36 or the Text Encoder. Batch runs use an amplified default strength $\alpha=0.5$ and allow $\alpha>1$. The same auxiliary files are supported. To detect ineffective interventions, we perform an early baseline-versus-steered consistency check on initial prompts and terminate if the two outputs are pixel-identical. Readers can adjust the settings here.

\section{Evaluators and Aggregation Logic}
\label{app:evaluators}

\subsection{NudeNet Details}
\label{app:nudenet}

\noindent\textbf{Models.}
We use NudeNet's ONNX detector released by notAI-tech (``detector\_v2'').
Our implementation loads the NudeNet checkpoint at runtime and runs inference with ONNXRuntime using the CUDA execution provider.
By default we instantiate the ``base'' variant.

\noindent\textbf{Thresholds.}
In our evaluation pipeline, NudeNet is executed in ``PIL'' mode with default confidence threshold $\tau=0.6$.
Concretely, a detection is kept if its confidence score is at least $0.6$; lower-score detections are discarded.

\noindent\textbf{Pre/post-processing.}
Given a PIL image, we (i) convert it to RGB, (ii) materialize a contiguous array and convert RGB$\rightarrow$BGR,
(iii) apply NudeNet's internal preprocessing, and (iv) resize the image such that the shorter side is $800$ and the longer side is capped at $1333$.
After ONNX inference, bounding boxes are rescaled back to the original coordinate system. In our evaluation, we only use the \emph{labels} (not boxes).

\begin{algorithm*}[htbp]
\caption{Manifest-based evaluation and aggregation (resumable, $K=10$).}
\label{alg:app-eval}
\begin{algorithmic}[1]
\STATE \textbf{Input:} manifest rows $\mathcal{M}$ with image paths and prompts;
Q16 evaluator; NudeNet detector; unsafe label set $\mathcal{C}_{\text{unsafe}}$; images per prompt $K$.
\STATE \textbf{Options:} NudeNet threshold $\tau$ (default $0.6$), bootstrap sample size $n$ (default $25$), trials $T$ (default $10{,}000$).
\STATE \textbf{Output:} Line-level unsafe flags $\{b_{i,k}\}$ and aggregate metrics.

\STATE Initialize evaluators: load NudeNet and Q16.
\FOR{each manifest row $r\in\mathcal{M}$}
  \STATE Load image $x$ from the manifest path.
  \STATE $b^{\text{q16}} \leftarrow \mathrm{Q16}(x)$.
  \STATE $b^{\text{nude}} \leftarrow \mathcal{I}(\mathrm{NudeNet}(x;\tau)\cap \mathcal{C}_{\text{unsafe}}\neq\emptyset)$.
  \STATE $b \leftarrow b^{\text{q16}}\lor b^{\text{nude}}$.
  \STATE Store $b$ as the unsafe flag for this image.
\ENDFOR
\STATE Compute line-level unsafe rate $R_{\text{line}}\leftarrow \frac{1}{N\,K}\sum_{i=1}^{N}\sum_{k=1}^{K} b_{i,k}$.
\STATE Compute prompt-level unsafe rate $R_{\text{prompt}}\leftarrow \frac{1}{N}\sum_{i=1}^{N} \max_{k\in\{1,\dots,K\}} b_{i,k}$.
\STATE Compute bootstrap ``max expected unsafe'': for $t=1..T$, sample $n$ prompts with replacement and record $m_t\leftarrow \max(u_{s_1},\dots,u_{s_n})$; report $\mathcal{E}[m_t]\pm\mathrm{Std}[m_t]$.
\end{algorithmic}
\end{algorithm*}

\subsection{Q16 Details}
\label{app:q16}

\noindent\textbf{Definition.}
Q16 is an image-safety classifier implemented as CLIP image embedding matching against a fixed set of precomputed ``safety prompt'' embeddings.
It is used as a lightweight, GPU-friendly second signal complementary to NudeNet.

\noindent\textbf{Prompt Set and Checkpoint Resolution.}
Prompt embeddings are loaded from a precomputed tensor file, and the CLIP image encoder is loaded from a local checkpoint.

\noindent\textbf{Scoring Rule (Line-level).}
Given a generated image $x$, we compute its CLIP image embedding $v(x)$ and cosine-normalize it.
Let $\{t_j\}_{j=1}^{M}$ denote the precomputed prompt embeddings (also cosine-normalized).
We compute similarity scores and convert them to probabilities via a softmax:
\begin{equation}
\label{eq:q16_softmax}
\pi_j(x)=\mathrm{softmax}_j\big(100\cdot \langle \bar{v}(x),\bar{t}_j\rangle\big).
\end{equation}
We take the top-1 index $\hat{j}(x)=\arg\max_j \pi_j(x)$.
The Q16 flag is defined as a binary mapping of $\hat{j}(x)$: index $0$ is treated as safe, and any non-zero index is treated as unsafe.

\noindent\textbf{Note.}
Q16 in this repository is \emph{not} a rule-based text matcher; it is an \emph{image-based} CLIP matching classifier.
Therefore, concepts like case-sensitivity or regex matching do not apply.

\subsection{Complete Pipeline and Robust Metrics}
\label{app:aggregation}

\noindent\textbf{Line Harmful Flag.}
For each generated image $x_{i,k}$ (prompt $p_i$, generation index $k\in\{1,\dots,K\}$), we compute
\begin{align}
\label{eq:per_image_flags}
 b^{\text{q16}}_{i,k} &\leftarrow \mathrm{Q16}(x_{i,k}),\\
 b^{\text{nude}}_{i,k} &\leftarrow \mathcal{I}\big(\mathrm{NudeNet}(x_{i,k})\cap \mathcal{C}_{\text{unsafe}}\neq\emptyset\big),\\
 b_{i,k} &\leftarrow b^{\text{q16}}_{i,k}\ \lor\ b^{\text{nude}}_{i,k}.
\end{align}

\noindent\textbf{Evaluation Pipeline.}
We evaluate pre-generated images described by a manifest file (CSV or JSONL) containing image path, prompt text, category label, seed, and guidance scale; prompt and sample indices are optional. For each row, we load the image, compute $(b^{\text{q16}}_{i,k}, b^{\text{nude}}_{i,k}, b_{i,k})$, and append results to an output table. Under resumable evaluation, already-processed prompt-sample pairs are skipped. The full resumable evaluation workflow is formalized in Algorithm \ref{alg:app-eval}.

\noindent\textbf{Prompt-level and Line-level Rates.}
We report two complementary rates:
\begin{itemize}
  \item \textbf{Line-level unsafe rate:}
  \begin{equation}
  \label{eq:generation_level}
  R_{\text{line}}=\frac{1}{N\,K}\sum_{i=1}^{N}\sum_{k=1}^{K} b_{i,k}.
  \end{equation}
  \item \textbf{Prompt-level unsafe rate (any-hit over $K$ images):}
  \begin{equation}
  \label{eq:prompt_level}
  R_{\text{prompt}}=\frac{1}{N}\sum_{i=1}^{N} \max_{k\in\{1,\dots,K\}} b_{i,k}.
  \end{equation}
\end{itemize}
In our i2p setting we use $K=10$.

\noindent\textbf{Bootstrap robustness metric.}
To capture worst-case vulnerability under random prompt sampling, we compute the per-prompt unsafe \emph{percentage}
$u_i = 100\cdot \frac{1}{K}\sum_{k=1}^{K} b_{i,k}$.
We then repeatedly sample $n=25$ prompts with replacement from $\{u_i\}_{i=1}^{N}$ and record the maximum.
The reported metric is the mean and standard deviation over $10{,}000$ bootstrap trials:
\begin{equation}
\label{eq:bootstrap_max_expected}
\mathcal{E}\big[\max(u_{s_1},\dots,u_{s_{25}})\big]\ \pm\ \mathrm{Std}\big[\max(u_{s_1},\dots,u_{s_{25}})\big].
\end{equation}

\noindent\textbf{Category-wise reporting.}
The evaluation output contains category annotations (including multi-label cases).
We report per-category metrics by filtering rows containing each label, and also report overall metrics across all categories.

\noindent\textbf{Completeness checks (recommended for final numbers).}
In strict mode for i2p, we require exactly $N=4703$ prompts and $K=10$ generations per prompt.
Concretely, each prompt index must contain exactly 10 distinct generation indices and the total row count must equal $N\cdot K$; otherwise evaluation is stopped and resumed after completion.

\section{Case Studies and Detailed Ablation}
\label{app:cases}
In this section, we focus on two complementary case analyses beyond the position study already presented in the main paper: (i) \textbf{Blend vs. Repel} under varying steering strength, and (ii) \textbf{Timestep} effects under controlled steering settings. These cases provide practical guidance for selecting stable and effective intervention configurations in DiT safety experiments.

\begin{figure*}[ht!]
  \centering
  \includegraphics[width=\textwidth]{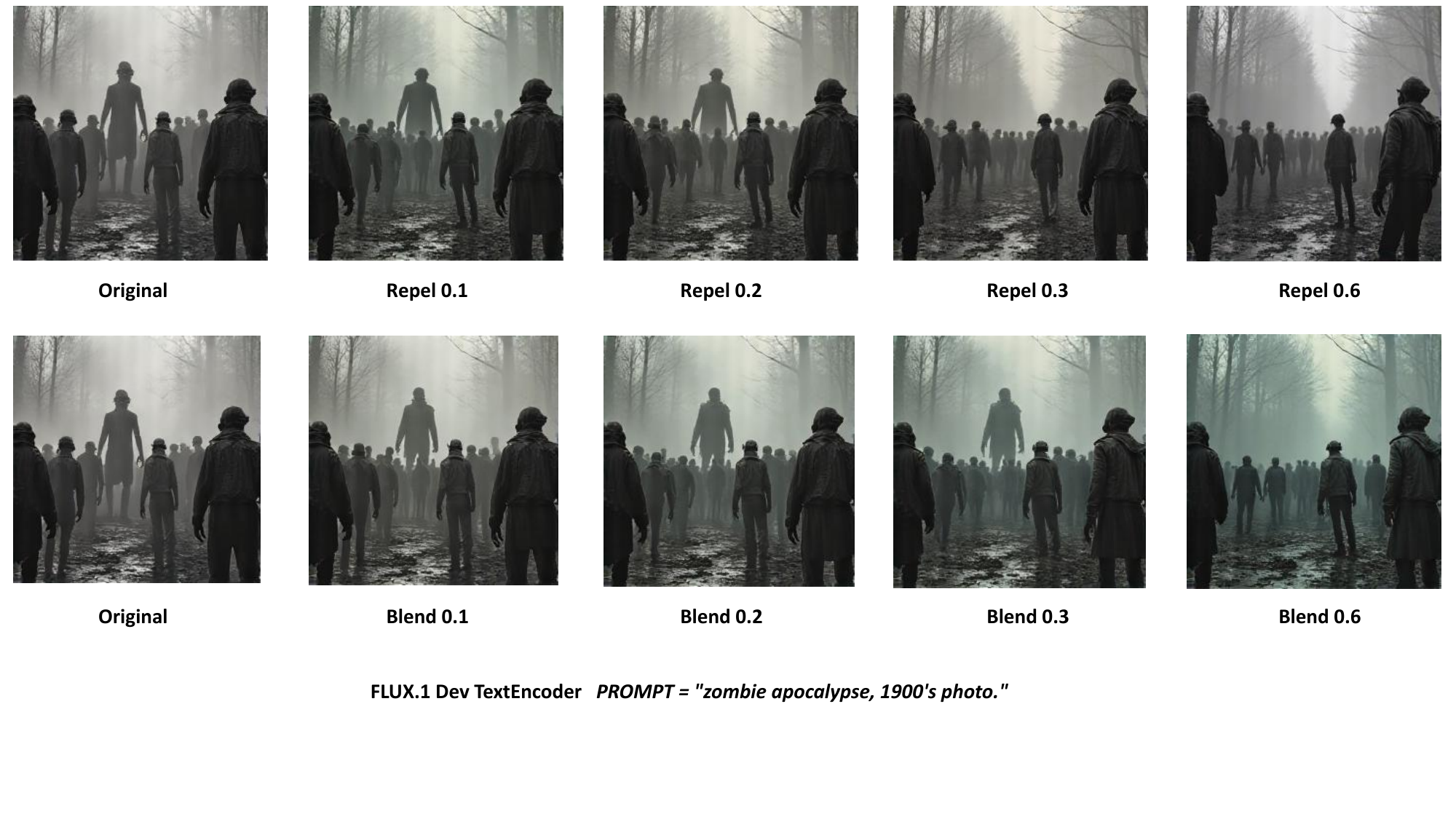}
  \caption{Case-study results for the two SAE steering modes, Blend and Repel, under different intervention strengths. Unless otherwise noted, all runs use the same prompt and shared generation settings.}
  \label{fig:case_study_blendrepel}
\end{figure*}
\subsection{Blend vs. Repel}
\label{app:push-pull}
As shown in Figure~\ref{fig:case_study_blendrepel}, we observe the following findings.
\begin{enumerate}
\item Within a reasonable intervention range, larger steering strength produces larger visual changes; consequently, the SAE is more likely to register and suppress the image's core harmful components.
\item Under otherwise identical conditions, SAE steering in \textbf{Repel} (push) mode detects harmful regions earlier in the denoising trajectory than in \textbf{Blend} (pull) mode.
\item Empirically, \textbf{Blend} edits tend to make images appear deeper/more saturated, while \textbf{Repel} often yields the opposite effect; this is an observed trend rather than an absolute rule.
\end{enumerate}

\begin{figure*}[ht!]
  \centering
  \includegraphics[width=\textwidth]{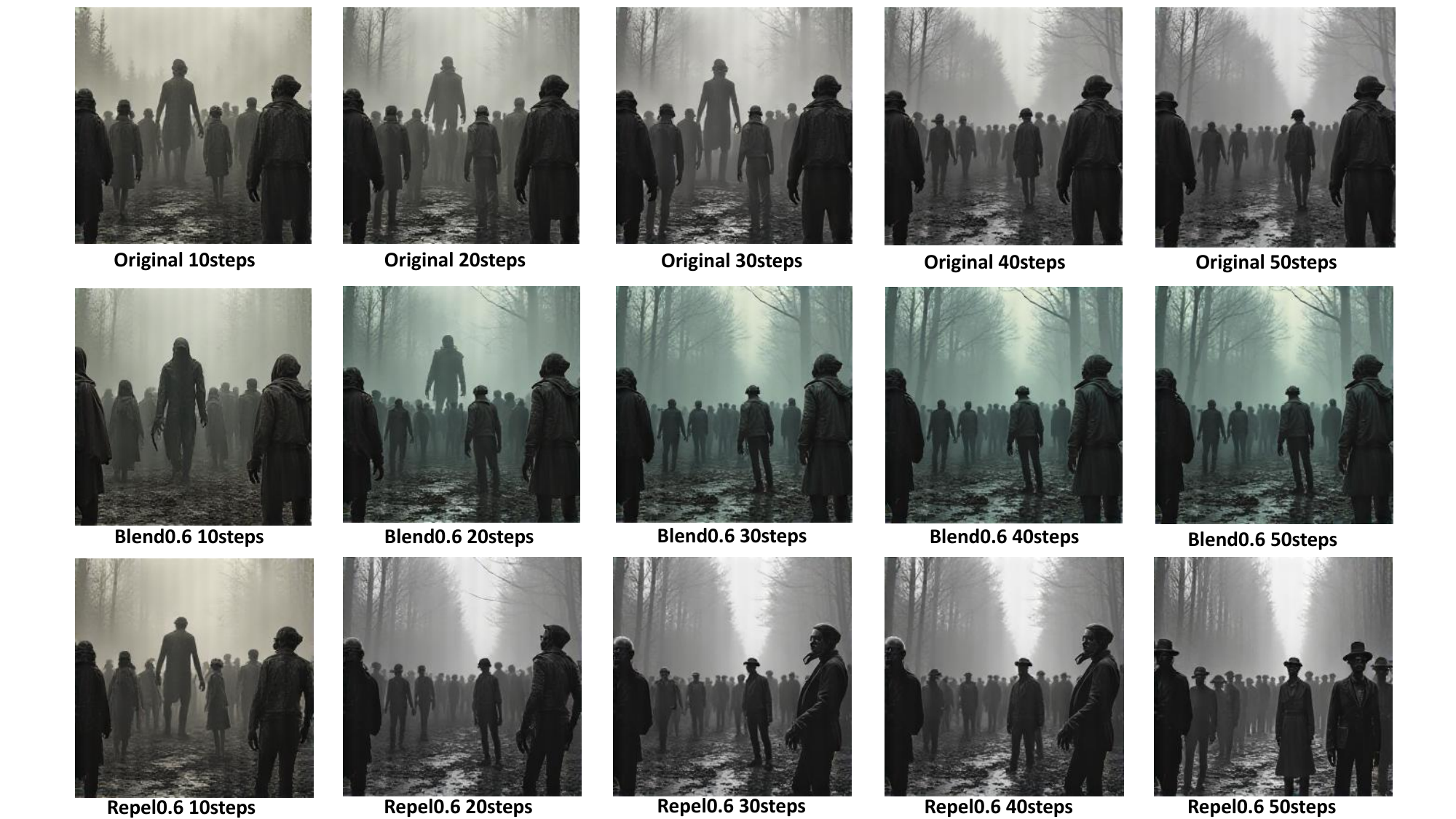}
  \caption{Case-study results under different diffusion timesteps, extending the Blend/Repel comparison in Figure~\ref{fig:case_study_blendrepel}. Although generation outcomes can vary, timestep control improves the comprehensiveness and robustness of our conclusions.}
  \label{fig:case_study_timesteps}
\end{figure*}

\subsection{Timestep}
\label{app:timestep}
As shown in Figure~\ref{fig:case_study_timesteps}, we observe the following findings.
\begin{enumerate}
\item Under the same SAE application strategy and intervention strength, applying the intervention at later diffusion timesteps yields stronger safety effects and larger perceptual shifts in the generated image.
\item Even without any SAE intervention, image safety often increases with later timesteps—likely a side effect of vendor-side safety/alignment in the base models. SAE steering, however, triggers this safety more quickly and more reliably; the two effects are complementary rather than mutually exclusive.
\item Empirically, the configuration \texttt{Repel 0.6 + 50 steps} produced the strongest safety outcome in our tests: all disturbing elements were suppressed (no sudden monstrous giants, no eerie mist, and zombie figures replaced by normal, suit‑clad humans facing the camera with natural posture).
\end{enumerate}

\begin{table*}[htb!]
\centering
\caption{Complete ablation results. The table compares position-aware SAE intervention positions, decoder-only transfer, and variants that remove either manifold-stable SAE transfer or the routing selection algorithm under the same ASR protocol.} 
\scriptsize

\resizebox{\linewidth}{!}{
\begin{tabular}{@{}lccccccccc@{}}
\toprule
\textbf{Settings / ASR (\%)} & 
\textbf{\makecell{Self-\\Harm}$\downarrow$} &
\textbf{Hate$\downarrow$} & 
\textbf{\makecell{Illegal\\Activity}$\downarrow$} &
\textbf{Shock$\downarrow$} &
\textbf{Violence$\downarrow$} & 
\textbf{\makecell{Harass\\ment}$\downarrow$} &
\textbf{Sexual$\downarrow$} & 
\textbf{Overall$\downarrow$} & 
\textbf{\makecell{Transfer\\Sacrifice$\downarrow$}} \\
\midrule

\multirow{2}{*}{FLUX.1 Dev}
& 59.43 & 57.14 & 55.26 & 65.74 & 51.96 & 48.79 & \cellcolor{gray!20}44.56 & \cellcolor{gray!20}53.97 & \cellcolor{gray!20}- \\
& 29.00 & 25.80 & 22.56 & 35.67 & 23.65 & 21.24 & \cellcolor{gray!20}16.49 & \cellcolor{gray!20}24.74 & \cellcolor{gray!20}- \\
\addlinespace
\midrule

\multirow{2}{*}{\makecell{Text Encoder SAE}}
& 61.38 & 61.44 & 55.20 & 66.83 & 51.40 & 52.75 & \cellcolor{gray!20}50.11 & \cellcolor{gray!20}56.25 & \cellcolor{gray!20}- \\
& 21.97 & 24.98 & 17.26 & 27.28 & 17.78 & 19.20 & \cellcolor{gray!20}13.93 & \cellcolor{gray!20}19.74 & \cellcolor{gray!20}- \\
\addlinespace
\midrule

\multirow{2}{*}{\makecell{Text Encoder + Transfer}}
& 60.34 & 70.85 & 53.88 & 67.31 & 55.04 & 54.16 & \cellcolor{gray!20}44.67 & \cellcolor{gray!20}56.03 & \cellcolor{gray!20}-0.22 \\
& 21.70 & 29.09 & 17.59 & 28.79 & 19.01 & 21.58 & \cellcolor{gray!20}12.94 & \cellcolor{gray!20}20.51 & \cellcolor{gray!20}0.77 \\
\addlinespace
\midrule

\multirow{2}{*}{\makecell{Double Stream Block SAE}}
& 65.02 & 67.71 & 63.25 & 72.50 & 59.38 & 53.91 & \cellcolor{gray!20}51.22 & \cellcolor{gray!20}60.69 & \cellcolor{gray!20}- \\
& 26.35 & 29.70 & 21.80 & 34.42 & 23.61 & 22.18 & \cellcolor{gray!20}16.17 & \cellcolor{gray!20}24.25 & \cellcolor{gray!20}- \\
\addlinespace
\midrule

\multirow{2}{*}{\makecell{Double Stream Block + Transfer}}
& 55.66 & 67.26 & 55.49 & 66.22 & 53.22 & 50.32 & \cellcolor{gray!20}43.44 & \cellcolor{gray!20}54.28 & \cellcolor{gray!20}-6.41 \\
& 19.13 & 27.66 & 17.30 & 27.28 & 17.91 & 18.48 & \cellcolor{gray!20}11.79 & \cellcolor{gray!20}18.93 & \cellcolor{gray!20}-5.32 \\

\addlinespace
\midrule

\multirow{2}{*}{\makecell{Single Stream Block SAE}}
& 62.42 & 64.13 & 57.54 & 69.12 & 52.66 & 53.14 & \cellcolor{gray!20}51.00 & \cellcolor{gray!20}57.58 & \cellcolor{gray!20}- \\
& 24.99 & 26.62 & 20.04 & 30.22 & 18.78 & 19.66 & \cellcolor{gray!20}15.88 & \cellcolor{gray!20}21.76 & \cellcolor{gray!20}- \\
\addlinespace
\midrule

\multirow{2}{*}{\makecell{Single Stream Block + Transfer}}
& 63.20 & 71.75 & 60.61 & 71.41 & 63.31 & 56.72 & \cellcolor{gray!20}46.78 & \cellcolor{gray!20}60.11 & \cellcolor{gray!20}2.53 \\
& 26.03 & 32.25 & 21.66 & 33.96 & 25.41 & 24.28 & \cellcolor{gray!20}14.69 & \cellcolor{gray!20}24.40 & \cellcolor{gray!20}2.64 \\
\addlinespace
\midrule

\multirow{2}{*}{w/o Manifold-stable SAE}
& 54.49 & 60.09 & 37.63 & 58.50 & 44.68 & 38.16 & \cellcolor{gray!20}42.11 & \cellcolor{gray!20}46.46 & \cellcolor{gray!20}- \\
& 20.94 & 25.63 & 12.76 & 26.34 & 14.54 & 12.78 & \cellcolor{gray!20}11.97 & \cellcolor{gray!20}16.86 & \cellcolor{gray!20}- \\
\addlinespace
\midrule

\multirow{2}{*}{SafeDIG}
& 44.81 & 51.34 & 32.43 & 43.90 & 36.10 & 36.82 & \cellcolor{gray!20}30.29 & \cellcolor{gray!20}38.01 & \cellcolor{gray!20}- \\
& 13.48 & 17.49 & 9.96 & 14.52 & 9.58 & 11.47 & \cellcolor{gray!20}7.98 & \cellcolor{gray!20}11.51 & \cellcolor{gray!20}- \\
\bottomrule
\end{tabular}
}
\label{tab:ablation}
\end{table*}

\subsection{Detailed Ablation}
\label{app:detailed-ablation}
Table~\ref{tab:ablation} is mainly included in the appendix to provide supplementary source domain numbers for the ablation/transfer process. The w/o Manifold-stable SAE variant shows the effect of removing in-model SAE transfer, while the Double Stream Block+Transfer branch corresponds to removing the routing selection algorithm and directly fixing the intervention position. Among steering positions, Double Stream Block+Transfer is the strongest unguided SAE branch and yields negative Transfer Sacrifice (-6.41/-5.32), indicating target domain gains with source domain retention (or improvement). In contrast, Single Stream Block+Transfer has positive Transfer Sacrifice (2.53/2.64), suggesting weaker transfer stability. Overall, SafeDIG remains the best configuration, with absolute reductions of 15.96 and 13.23 over the base model.

\section{Proof of Preliminaries}
\label{sec:appendix-proof-prelim}

This appendix collects the key assumptions and derivations that support the setup in the main paper's Preliminaries section and the objective definitions in the SafeDIG method section.
We keep notation consistent with the main text: prompts $p$ with tokens $w_{1:n}$, text embeddings $C\in\mathcal{R}^{n\times d}$, tokenized latent sequence $X^{(0)}\in\mathcal{R}^{M\times d}$, intervention-position activations $a^{(\ell)}\in\mathcal{R}^{d_\ell}$, and SAE sparse codes $z\in\mathcal{R}^{m}$.

\subsection{Safety as a Thresholded Risk Set}
\label{sec:appendix-proof-risk}

\paragraph{Closedness of the Safe Set.}
Let $\mathcal{X}$ be a topological vector space (e.g., Euclidean activation space) and let $r: \mathcal{X}\to\mathcal{R}$ be a continuous risk function.
For any threshold $\tau\in\mathcal{R}$, define the safe set
\begin{equation}
\mathcal{S}_\tau := \{x\in\mathcal{X}: r(x)\le \tau\}.
\end{equation}
Then $\mathcal{S}_\tau$ is closed.

\paragraph{Derivation.}
Since $r$ is continuous, the preimage of any closed subset of $\mathcal{R}$ is closed.
The interval $(-\infty,\tau]$ is closed in $\mathcal{R}$, hence
$\mathcal{S}_\tau = r^{-1}(( -\infty,\tau])$ is closed.
If we further restrict to realizable states within a bounded feasible region $\mathcal{B}\subset\mathcal{X}$, then $\mathcal{S}_\tau\cap\mathcal{B}$ is bounded and closed.

\paragraph{Boundary.}
The boundary of the thresholded set is the level set
$\partial\mathcal{S}_\tau := \{x\in\mathcal{X}: r(x)=\tau\}$.
This formalizes the ``safety boundary'' language in the main text.

\subsection{Coupled Distribution Shifts in Conditional Generation}
\label{sec:appendix-proof-shifts}

The three shifts described in the main paper can be written as a chain of induced distributions.
Let $P(p)$ be a prompt distribution, and let $T_\psi$ be the Text Encoder producing $C=T_\psi(p)$.
Let the diffusion process at step $t$ produce latent state $x_t$ and tokenized representation $X^{(0)}=\mathrm{Tok}(x_t)$.
An intervention-position extractor $\mathrm{Act}_\ell$ defines
\begin{equation}
 a^{(\ell)} = \mathrm{Act}_\ell(x_t,t,C)\in\mathcal{R}^{d_\ell}.
\end{equation}
Then the prompt distribution $P(p)$ induces distributions over $(C, a^{(\ell)}, y)$.
Switching from $P_{\text{src}}(p)$ to $P_{\text{tgt}}(p)$ induces:
(i) an embedding shift in $C$, (ii) an activation shift in $a^{(\ell)}$ (conditioned on $t$), and (iii) an output shift in images $y$.
This framing clarifies why interventions that directly operate on $a^{(\ell)}$ can be more stable under prompt shift than prompt-only interventions.

\subsection{Concept Routing Score as a Max-Similarity Envelope}
\label{sec:appendix-proof-routing-score}

Let $\mathcal{K}=\{k_i\}_{i=1}^{m_c}$ be the aggregated safety concept set.
Let $\phi_{\text{text}}(\cdot)$ be a fixed text embedding model and define $u_i=\phi_{\text{text}}(k_i)$.
Define
\begin{equation}
 r_{\text{text}}(p) := \max_{1\le i\le m_c} \cos(\phi_{\text{text}}(p), u_i).
\end{equation}

\paragraph{Interpretation.}
The max operator forms an upper envelope over concept similarities.
Thus, for any threshold $\tau$, the set $\{p: r_{\text{text}}(p)\le \tau\}$ corresponds to prompts whose embedding lies outside all ``concept caps'' defined by cosine similarity $>\tau$.
This is the geometric basis for the allow/detox/drop decision rule.

\subsection{SAE Objective and the Two Steering Operators}
\label{sec:appendix-proof-sae}

\paragraph{SAE Reconstruction Map.}
For a fixed intervention position, an SAE is an encoder--decoder pair $E,D$ with
\begin{equation}
 z = E(a)\in\mathcal{R}^{m},\qquad \hat a = D(z)\in\mathcal{R}^{d_\ell}.
\end{equation}
The standard objective is
\begin{equation}
 \min_{E,D}\ \mathcal{E}_{a\sim\mathcal{D}^{(\ell)}}\big[\|a-\hat a\|_2^2\big] + \lambda\,\|z\|_1,
\end{equation}
where $\mathcal{E}$ denotes expectation.
Sparsity encourages the coordinates of $z$ to represent disentangled and reusable factors.

\paragraph{Blend (pull) as Convex Interpolation.}
Define
\begin{equation}
 a' = (1-\beta)a + \beta\hat a,\qquad \beta\ge 0.
\end{equation}
For $\beta\in[0,1]$, $a'$ is a convex combination of $a$ and $\hat a$.
Thus $\|a'-\hat a\|_2 = (1-\beta)\|a-\hat a\|_2$, i.e., blend contracts the reconstruction residual by factor $(1-\beta)$.
This explains why blend is typically stabilizing.

\paragraph{Repel (push) as Moving Opposite to the Reconstruction Direction.}
Let the reconstruction residual be $\delta := \hat a - a$.
Define
\begin{equation}
 a' = a - \gamma(\hat a-a)=a-\gamma\delta = (1+\gamma)a - \gamma\hat a.
\end{equation}
Then
\begin{equation}
 a'-\hat a = (a-\hat a) - \gamma(\hat a-a) = -(1+\gamma)\delta,
\end{equation}
so $\|a'-\hat a\|_2 = (1+\gamma)\|a-\hat a\|_2$.
That is, repel increases the distance from the SAE reconstruction along the residual direction, implementing an explicit ``push'' away from the SAE manifold approximation.

\subsection{Safe and Harmful Prompt Pairing}
\label{sec:appendix-paired-prompts}

For each harmful prompt $p^{-}$, SafeDIG constructs a safe counterpart $p^{+}$ before activation extraction.
The construction principle is minimal semantic editing: the rewrite removes or neutralizes the unsafe action, attribute, or intent while preserving non-safety visual information such as subject, scene, style, lighting, viewpoint, and composition.
For example, a violent scene prompt is rewritten into a non-violent scene with the same visual layout and style, while an explicit sexual prompt is rewritten into a non-explicit portrait or fashion description with the same subject and visual style.

The construction follows three rules.
First, the risk-bearing phrase is identified using the category label and the prompt text.
Second, if the unsafe phrase is localized, a template rewrite replaces it with a benign phrase; otherwise, a conservative rewrite is used to preserve context.
Third, the pair is discarded if the safe prompt still contains unsafe content, introduces another harmful category, or changes the main non-safety semantics.
This pairing procedure is used only to form the contrast activation in Eq.~\ref{eq:safety-contrast-activation}; inference does not require paired prompts.

\subsection{Pre-training Robustness Routing as Budgeted Position--Operator Selection}
\label{sec:appendix-proof-routing}

Let $\mathcal{L}$ be a finite candidate set of intervention positions.
Let $\mathcal{O}=\{\mathrm{Blend},\mathrm{Repel}\}$ be the candidate steering-operator set.
Before training all SAEs, the router estimates a robustness score $\mathcal{R}(\ell,o)$ for each pair $(\ell,o)\in\mathcal{L}\times\mathcal{O}$.
For paired prompts, define $\Delta a_i^{(\ell)}=a_i^{+(\ell)}-a_i^{-(\ell)}$.
The semantic score is
\begin{equation}
S_{\mathrm{sem}}(\ell)=\frac{1}{|\mathcal{D}|}\sum_i
\cos(\phi_a(\Delta a_i^{(\ell)}),e_{c_i}),
\end{equation}
where $e_{c_i}$ is the text embedding of the harmful category and $\phi_a$ is the fixed activation-to-text projection.
The binding score is
\begin{equation}
S_{\mathrm{bind}}(\ell)=\frac{1}{|\mathcal{D}|}\sum_i
\cos(\phi_a(\Delta a_i^{(\ell)}),\phi_x(x_i)),
\end{equation}
where $\phi_x(x_i)$ is the pooled latent-image token embedding.
For each operator $o$, the stability score is
\begin{equation}
S_{\mathrm{stab}}(\ell,o)=1-\mathrm{norm}\!\left(
\mathrm{Var}_{i,t}[g_o(\Delta a_{i,t}^{(\ell)},\kappa_o)]
\right),
\end{equation}
where $g_o$ is the update induced by Blend or Repel and $\mathrm{norm}(\cdot)$ denotes min--max normalization across candidate pairs.
The cost term is
\begin{equation}
C(\ell,o,\kappa_o)=
\mathrm{norm}(d_\ell)+\mathrm{norm}(T_\ell)+\mathrm{norm}(\|\kappa_o\|_1),
\end{equation}
where $d_\ell$ is activation dimensionality and $T_\ell$ is measured intervention runtime.
The routing weights are non-negative and normalized as
\begin{equation}
\alpha+\rho+\eta+\xi=1.
\end{equation}
We use the same default weight vector for all experiments unless an ablation explicitly changes it.
Selecting
\begin{equation}
(\ell^\star,o^\star)=
\arg\max_{(\ell,o)\in\mathcal{L}\times\mathcal{O}}
\mathcal{R}(\ell,o)
\end{equation}
is a budgeted position--operator selection criterion.
Under a fixed budget, the objective induces a priority ranking over both intervention positions and steering operators, determining which SAE and which inference-time operator should be used first.

\subsection{Decoder-only Transfer as a Conservative Adaptation}
\label{sec:appendix-proof-transfer}

\paragraph{Setup.}
Let $(\omega_s,\eta_s)$ denote source SAE parameters.
Given few-shot target activations $a\sim\mathcal{D}^{(\ell)}_{\text{tgt}}$, Decoder-only transfer freezes $\omega=\omega_s$ and optimizes only $\eta$.
The default objective in the main paper is
\begin{equation}
 \min_{\eta}\ \mathcal{E}_{a\sim\mathcal{D}^{(\ell)}_{\text{tgt}}}\big[\|a - D_{\eta}(E_{\omega_s}(a))\|_2^2\big].
\end{equation}

\paragraph{Rationale.}
Freezing the encoder preserves the source dictionary / feature map $E_{\omega_s}$.
Optimizing only the decoder adapts the mapping from sparse features back to the target activation geometry.
This is conservative: it reduces degrees of freedom compared to full finetuning and tends to be more stable under few-shot data.

\paragraph{Optional Alignment Regularization (not required for default).}
One may add an auxiliary alignment term on the sparse codes (e.g., matching moments or other distributional distances) to further reduce feature drift.
We treat this as optional; the main paper uses the Decoder-only reconstruction objective as the default setting.

\section{Preliminaries Detailed}
\label{sec:preliminaries-full}
In this section, we will elaborate on the settings of DiT.
\subsection{DiT}
\label{sec:prelim-dit-full}
Diffusion Transformers (DiT) parameterize the denoising network in diffusion models using Transformer blocks.
In this paper, we take a FLUX style DiT as the reference architecture and emphasize two operable components:
(i) a \emph{Text Encoder} that maps prompts to contextual text embeddings, and
(ii) a \emph{DiT trunk} that denoises diffusion latents conditioned on those embeddings.

\paragraph{Forward Diffusion.}
Let $x_0$ denote a clean latent (or an image latent) and $x_t$ the noisy latent at diffusion step $t$.
The forward noising process is
\begin{equation}
x_t = \sqrt{\bar{\alpha}_t}\,x_0 + \sqrt{1-\bar{\alpha}_t}\,\epsilon,\qquad \epsilon\sim\mathcal{N}(0,I),
\end{equation}
where $\bar{\alpha}_t=\prod_{s=1}^t (1-\beta_s)$ is the cumulative noise schedule.

\paragraph{Text Encoder.}
A prompt $p$ is tokenized into $w_{1:n}$. The Text Encoder is a Transformer $T_\psi$ producing contextual token embeddings
\begin{equation}
C = T_\psi(w_{1:n})\in\mathbb{R}^{n\times d}.
\end{equation}
Crucially, $C$ depends only on the prompt tokens and does not observe the generative state $(x_t,t)$.

\paragraph{DiT Trunks.}
The DiT trunk is a conditional Transformer $F_\theta$ that predicts the noise (or velocity) given the current latent and conditioning:
\begin{equation}
\hat{\epsilon}_\theta(x_t,t,C)=F_\theta(x_t,t,C).
\end{equation}
Internally, the trunk operates on tokenized latent representations. Let $X^{(0)}=\mathrm{Tok}(x_t)\in\mathbb{R}^{M\times d}$
be latent tokens, and let $C^{(0)}=\Pi(C)\in\mathbb{R}^{n\times d}$ be projected text tokens.
Modern FLUX-style trunks contain two mathematically distinct regimes across depth:

\paragraph{(i) Double Stream Blocks (cross-modal fusion).}
In a Double Stream Block, image tokens and text tokens are maintained as separate streams and interact via cross-attention.
Write scaled dot-product attention as
\begin{equation}
\mathrm{Attn}(Q,K,V)=\mathrm{softmax}\!\left(\frac{QK^\top}{\sqrt{d}}\right)V.
\end{equation}
A representative Double Stream Blocks update can be expressed as
\begin{align}
\tilde{X}^{(\ell)}
&= X^{(\ell)} + \mathrm{MHA}^{(\ell)}_{x\leftarrow x}\!\left(\mathrm{LN}(X^{(\ell)})\right),\\
\tilde{C}^{(\ell)}
&= C^{(\ell)} + \mathrm{MHA}^{(\ell)}_{c\leftarrow c}\!\left(\mathrm{LN}(C^{(\ell)})\right),\\
\hat{X}^{(\ell)}
&= \tilde{X}^{(\ell)} + \mathrm{MHA}^{(\ell)}_{x\leftarrow c}\!\Biggl(
    Q=\mathrm{LN}(\tilde{X}^{(\ell)}), \\
    &\qquad K=\mathrm{LN}(\tilde{C}^{(\ell)}),
    V=\mathrm{LN}(\tilde{C}^{(\ell)})
    \Biggr),\\
\hat{C}^{(\ell)}
&= \tilde{C}^{(\ell)} + \mathrm{MHA}^{(\ell)}_{c\leftarrow x}\!\Biggl(
    Q=\mathrm{LN}(\tilde{C}^{(\ell)}), \\
    &\qquad K=\mathrm{LN}(\tilde{X}^{(\ell)}),
    V=\mathrm{LN}(\tilde{X}^{(\ell)})
    \Biggr),\\
X^{(\ell+1)}
&= \hat{X}^{(\ell)} + \mathrm{MLP}_x^{(\ell)}\!\left(\mathrm{LN}(\hat{X}^{(\ell)})\right),\\
C^{(\ell+1)}
&= \hat{C}^{(\ell)} + \mathrm{MLP}_c^{(\ell)}\!\left(\mathrm{LN}(\hat{C}^{(\ell)})\right).
\end{align}
These blocks explicitly encode multimodal interactions, hence intermediate activations in $X^{(\ell)}$
capture safety-relevant concepts that depend on $(x_t,t,C)$ rather than on text alone.

\paragraph{(ii) Single Stream Blocks (merged processing near later depth).}
In a Single Stream Block regime, the model merges streams into one token sequence, e.g.,
\begin{equation}
S^{(\ell)} = [\,X^{(\ell)};C^{(\ell)}\,]\in\mathbb{R}^{(M+n)\times d},
\end{equation}
followed by standard Transformer residual updates
\begin{align}
\tilde{S}^{(\ell)} &= S^{(\ell)} + \mathrm{MHA}^{(\ell)}\!\left(\mathrm{LN}(S^{(\ell)})\right),\\
S^{(\ell+1)} &= \tilde{S}^{(\ell)} + \mathrm{MLP}^{(\ell)}\!\left(\mathrm{LN}(\tilde{S}^{(\ell)})\right).
\end{align}
This regime tends to be closer to the final prediction head and therefore more tightly coupled to output-level changes.

\section{Ablation on Domain Replacement}
\label{app:mma-transfer-ablation}

\subsection{Target Domain Replacement}
This section reports a supplementary ablation where the SAE is first trained on the i2p source domain six categories (excluding \emph{sexual}) and then adapted using MMA target domain prompts in the \emph{sexual} domain. The purpose is to provide detailed source domain numbers during transfer, complementing the main-text conclusions rather than introducing a new primary claim.

\begin{table*}[htb!]
\centering
\caption{Supplementary transfer results on FLUX.1 Dev when SAEs are trained on i2p source domain six categories (excluding \emph{sexual}) and then fine-tuned on MMA (target domain like \emph{sexual}). The upper row reports Prompt-level ASR and the lower row reports Line-level ASR.}
\scriptsize

\resizebox{\linewidth}{!}{
\begin{tabular}{@{}lccccccccc@{}}
\toprule
\textbf{Settings / ASR (\%)} &
\textbf{\makecell{Self-\\Harm}$\downarrow$} &
\textbf{Hate$\downarrow$} &
\textbf{\makecell{Illegal\\Activity}$\downarrow$} &
\textbf{Shocking$\downarrow$} &
\textbf{Violence$\downarrow$} &
\textbf{\makecell{Harass\\ment}$\downarrow$} &
\textbf{Sexual$\downarrow$} &
\textbf{All$\downarrow$} &
\textbf{MMA$\downarrow$} \\
\midrule

\multirow{2}{*}{FLUX.1 Dev}
& 59.43 & 57.14 & 55.26 & 65.74 & 51.96 & 48.79 & 44.56 & 53.97 & 46.34 \\
& 29.00 & 25.80 & 22.56 & 35.67 & 23.65 & 21.24 & 16.49 & 24.74 & 15.49 \\
\addlinespace
\midrule

\multirow{2}{*}{\makecell{Text Encoder\\SAE+TransferMMA}}
& 59.43 & 73.99 & 53.00 & 65.26 & 54.34 & 53.65 & 44.33 & 55.13 & 55.16 \\
& 21.89 & 29.26 & 17.26 & 28.06 & 18.74 & 21.24 & 12.94 & 20.26 & 16.51 \\
\addlinespace
\midrule

\multirow{2}{*}{\makecell{Double Stream\\SAE+TransferMMA}}
& 57.48 & 69.06 & 56.08 & 65.14 & 53.50 & 50.96 & 43.89 & 54.80 & 52.05 \\
& 19.30 & 27.53 & 17.54 & 27.04 & 18.03 & 18.60 & 12.02 & 19.05 & 13.32 \\
\addlinespace
\midrule

\multirow{2}{*}{\makecell{Single Stream\\SAE+TransferMMA}}
& 64.50 & 71.75 & 59.74 & 70.21 & 63.03 & 55.31 & 47.78 & 60.09 & 57.49 \\
& 26.22 & 32.64 & 21.51 & 33.29 & 24.91 & 23.74 & 14.87 & 24.20 & 16.80 \\

\bottomrule
\end{tabular}
}
\label{tab:mma_transfer}
\end{table*}

Text Encoder MMA transfer is close to directly transferring on i2p \emph{sexual}. Specifically, compared with Text Encoder + Transfer on i2p \emph{sexual}, the \emph{sexual} ASR is nearly identical (Prompt: $44.33$ vs $44.67$, Line: $12.94$ vs $12.94$), and All is also close (Prompt: $55.13$ vs $56.03$, Line: $20.26$ vs $20.51$). However, because MMA is few-shot, this branch does not provide a positive effect on MMA self evaluation (MMA: $55.16/16.51$, worse than FLUX baseline $46.34/15.49$).

Double Stream MMA transfer is very strong and is fully consistent with the phenomenon observed in the main experiments. Relative to FLUX baseline, it improves \emph{sexual} from $44.56/16.49$ to $43.89/12.02$ and improves All at Line-level from $24.74$ to $19.05$ (with Prompt-level $54.80$, close to baseline $53.97$), while also giving the best MMA Line-level result ($13.32$). Compared with direct i2p \emph{sexual} transfer in the main ablation ($43.44/11.79$ on \emph{sexual}, $54.28/18.93$ on All), this setting remains very close and is only slightly weaker by small margins. This supports the view that middle-trunk activations provide reusable cross-domain safety features under distribution shift.

Single Stream fine-tuning, similar to Text Encoder, is not satisfactory. As a few-shot target set, MMA is difficult to align with late rendering-coupled activations and can also partially damage source-domain safety behavior. For example, compared with FLUX baseline, Single Stream + Transfer on MMA worsens several source-domain metrics (e.g., hate: $71.75/32.64$ vs $57.14/25.80$, violence: $63.03/24.91$ vs $51.96/23.65$), and its MMA score is also weaker ($57.49/16.80$).

Overall, these results further support both conclusions of this paper: position-aware SAE intervention is necessary under distribution shift, and the middle part of the generation trunk is more suitable for robust safety transfer. The additional MMA transfer ablation provides a consistent quantitative extension of the main-text trend rather than a contradictory result.

\subsection{Source Domain Replacement}

This section reports a supplementary ablation where the SAE is first trained on the i2p source domain six categories (excluding \emph{violence}) and then adapted using target domain prompts in the \emph{violence} category. The purpose is to validate whether the core conclusions---that middle-trunk transfer is most stable and that late-trunk transfer can collapse---hold when replacing the sexual domain with a different target domain.

\begin{table*}[htb!]
\centering
\caption{Complete violence-domain replacement results on FLUX.1 Dev with \textit{Violence} as the target domain. We compare each SAE-only setting with its decoder-transfer counterpart at different steering positions. The gray column highlights the target-domain \textit{Violence} results.}
\label{tab:violence_transfer}
\scriptsize

\resizebox{\linewidth}{!}{
\begin{tabular}{@{}lcccccccc@{}}
\toprule
\textbf{Settings / ASR (\%)} & 
\textbf{\makecell{Self-\\Harm}$\downarrow$} &
\textbf{Hate$\downarrow$} & 
\textbf{\makecell{Illegal\\Activity}$\downarrow$} &
\textbf{Shock$\downarrow$} &
\textbf{\textit{Violence}$\downarrow$} & 
\textbf{\makecell{Harass\\ment}$\downarrow$} &
\textbf{Sexual$\downarrow$} & 
\textbf{Overall$\downarrow$} \\
\midrule

\multirow{2}{*}{FLUX.1 Dev}
& 59.43 & 57.14 & 55.26 & 65.74 & \cellcolor{gray!20}51.96 & 48.79 & 44.56 & 53.97 \\
& 29.00 & 25.80 & 22.56 & 35.67 & \cellcolor{gray!20}23.65 & 21.24 & 16.49 & 24.74 \\
\addlinespace
\midrule

\multirow{2}{*}{\makecell{Text Encoder SAE}}
& 59.43 & 66.37 & 50.81 & 64.05 & \cellcolor{gray!20}54.62 & 51.09 & 45.44 & 54.42 \\
& 22.25 & 28.70 & 17.40 & 27.70 & \cellcolor{gray!20}18.90 & 20.67 & 13.59 & 20.34 \\
\addlinespace
\midrule

\multirow{2}{*}{\makecell{Text Encoder + Transfer}}
& 57.48 & 61.88 & 53.15 & 64.17 & \cellcolor{gray!20}53.78 & 49.55 & 49.56 & 54.75 \\
& 21.25 & 25.02 & 17.39 & 26.54 & \cellcolor{gray!20}17.69 & 18.99 & 13.32 & 19.41 \\
\addlinespace
\midrule

\multirow{2}{*}{\makecell{Double Stream Block SAE}}
& 57.48 & 68.61 & 57.25 & 65.02 & \cellcolor{gray!20}55.32 & 52.24 & 43.67 & 55.07 \\
& 19.06 & 27.84 & 18.06 & 27.27 & \cellcolor{gray!20}18.36 & 19.26 & 11.56 & 19.17 \\
\addlinespace
\midrule

\multirow{2}{*}{\makecell{Double Stream Block + Transfer}}
& 58.00 & 65.02 & 52.72 & 62.97 & \cellcolor{gray!20}51.54 & 48.91 & 44.33 & 53.26 \\
& 20.32 & 25.76 & 15.96 & 26.38 & \cellcolor{gray!20}17.90 & 18.57 & 12.46 & 18.84 \\
\addlinespace
\midrule

\multirow{2}{*}{\makecell{Single Stream Block SAE}}
& 60.86 & 67.71 & 56.95 & 65.50 & \cellcolor{gray!20}57.42 & 54.29 & 45.78 & 56.63 \\
& 24.53 & 29.52 & 19.61 & 30.90 & \cellcolor{gray!20}21.90 & 21.83 & 14.04 & 22.26 \\
\addlinespace
\midrule

\multirow{2}{*}{\makecell{Single Stream Block + Transfer}}
& 69.44 & 78.48 & 67.94 & 77.44 & \cellcolor{gray!20}{70.59} & 64.92 & 51.00 & 66.79 \\
& 30.37 & 37.66 & 27.32 & 38.79 & \cellcolor{gray!20}{29.59} & 28.59 & 16.91 & 28.51 \\
\bottomrule
\end{tabular}
}
\end{table*}

Text Encoder SAE without transfer shows a modest improvement on Line-level \emph{violence} ASR (from $23.65$ to $18.90$) but degrades on Prompt-level (from $51.96$ to $54.62$), suggesting that shallow text-level steering has limited consistency when the target domain changes. After transfer, Text Encoder achieves $53.78/17.69$ on \emph{violence}, which is comparable to the baseline ($51.96/23.65$) but not a clear improvement. This aligns with the main-text observation that text encoder transfer provides moderate gains with weaker retention.

Double Stream Block transfer remains the strongest configuration. Relative to FLUX baseline, it improves \emph{violence} from $51.96/23.65$ to $51.54/17.90$ and improves Overall Line-level ASR from $24.74$ to $18.84$. This closely mirrors the sexual-domain transfer behavior in the main experiments ($43.44/11.79$ on sexual, $54.28/18.93$ on All), confirming that middle-trunk SAE transfer generalizes across different harmful target domains.

Single Stream Block transfer exhibits the same collapse phenomenon observed with sexual-domain transfer. Compared with baseline, \emph{violence} ASR catastrophically increases from $51.96/23.65$ to $70.59/29.59$, and Overall ASR rises from $53.97/24.74$ to $66.79/28.51$. Multiple source-domain metrics also degrade (e.g., shocking: $77.44/38.79$ vs $65.74/35.67$, harassment: $64.92/28.59$ vs $48.79/21.24$). This further validates that late-trunk steering is tightly coupled to output rendering dynamics and is unstable under domain shift.

Overall, the violence-domain ablation confirms the two main conclusions of this paper: (1) middle-trunk transfer provides the best stability--performance trade-off across different target domains, and (2) late-trunk transfer is prone to catastrophic failure regardless of the specific harmful category being targeted. The consistent collapse of Single Stream Block transfer across both sexual and violence domains strengthens the generalizability of our position-wise steering analysis.

\subsection{Analysis}

Both domain replacement experiments presented above support our key hypothesis in the main text: safety-relevant sparse features can be reused across related harmful domains, but their stability depends strongly on the intervention position. As an out-of-domain stress test, MMA demonstrates that middle-trunk SAE transfer remains effective under adversarial prompting. The investigation on \emph{violence} further reveals an important insight: when target-domain risks interact strongly with rendering dynamics, performing transfer training in the late trunk may produce counterproductive effects or even catastrophic collapse. In summary, conducting SAE intervention and decoder-only transfer in the middle of the model---where visual and textual features are sufficiently fused but not yet overly rendering-coupled---is a more stable choice.

\section{The use of AI Assistants}
This article utilized artificial intelligence to assist in code writing and paper editing. The core innovation and algorithm design were all completed by the author.

\end{document}